\documentclass{article}

\PassOptionsToPackage{numbers, compress}{natbib}  

\usepackage[preprint]{neurips_2022}

\usepackage{url}            
\usepackage[utf8]{inputenc} 
\usepackage[T1]{fontenc}    
\usepackage{microtype}      
\usepackage{xcolor}         
\usepackage{xspace}
\usepackage{amsmath,amssymb,amsfonts,amsthm,dsfont,pifont,bm,bbm,mathrsfs,mathtools,nicefrac}
\usepackage{algorithm,algpseudocode,listings}
\usepackage{subfigure}
\usepackage{booktabs,multirow,adjustbox,diagbox,threeparttable}
\usepackage[pagebackref=false,breaklinks=true,colorlinks=true,citecolor=blue,bookmarks=false]{hyperref}
\usepackage{cleveref}  

\usepackage{bm}
\usepackage{wrapfig}

\definecolor{myMagenta}{rgb}{0.9,0,0.4}

\usepackage{algorithm}
\usepackage{listings}
\usepackage{etoolbox}
\makeatletter
\AfterEndEnvironment{algorithm}{\let\@algcomment\relax}
\AtEndEnvironment{algorithm}{\kern2pt\hrule\relax\vskip3pt\@algcomment}
\let\@algcomment\relax
\newcommand\algcomment[1]{\def\@algcomment{\footnotesize#1}}
\renewcommand\fs@ruled{\def\@fs@cfont{\bfseries}\let\@fs@capt\floatc@ruled
  \def\@fs@pre{\hrule height.8pt depth0pt \kern2pt}%
  \def\@fs@post{}%
  \def\@fs@mid{\kern2pt\hrule\kern2pt}%
  \let\@fs@iftopcapt\iftrue}
\makeatother
    
\crefname{section}{Sec.}{Secs.}
\Crefname{section}{Section}{Sections}
\crefname{table}{Tab.}{Tabs.}
\Crefname{table}{Table}{Tables}
\crefname{figure}{Fig.}{Figs.}
\Crefname{figure}{Figure}{Figures}
\crefname{equation}{Eq.}{Eqs.}
\Crefname{equation}{Equation}{Equations}
\crefname{theorem}{Theorem}{Theorems}
\crefname{lemma}{Lemma}{Lemmas}
\crefname{corollary}{Corollary}{Corollaries}
\crefname{algorithm}{Algorithm}{Algorithms}

\usepackage{apptools}
\crefname{subappendix}{\IfAppendix{Sec.}{appendix}}{\IfAppendix{Secs.}{appendices}s}

\newcommand{\tocite}[1]{\textcolor{red}{[TO CITE]}}

\newcommand{\ieno}{\textit{i}.\textit{e}.}

\newcommand{\etcno}{\textit{etc}}

\newcommand{\f}{\bm{f}}
\newcommand{\F}{\bm{F}}
\newcommand{\btheta}{\btheta}

\newcommand{\x}{\bm{x}}
\newcommand{\bA}{\bm{A}}
\newcommand{\bB}{\bm{B}}

\newcommand{\bJ}{\bm{J}}

\newcommand{\rank}{\mathrm{Rank}}

\newcommand{\bpsi}{\bm{\psi}}

\newcommand{\bW}{\bm{W}}
\newcommand{\bD}{\bm{D}}
\newcommand{\sL}{\mathcal{L}}

\newtheorem{theorem}{Theorem}
\newtheorem{lemma}{Lemma}

\newtheorem{corollary}{Corollary}

\protected\def\ignorethis#1\endignorethis{}
\let\endignorethis\relax
\def\TOCstop{\addtocontents{toc}{\ignorethis}}
\def\TOCstart{\addtocontents{toc}{\endignorethis}}

\title{Rank Diminishing in Deep Neural Networks}
\author{
Ruili Feng$^1$, Kecheng Zheng$^{2,1}$, Yukun Huang$^1$, Deli Zhao$^{2,3}$, \\
\textbf{Michael Jordan}$^4$\textbf{,} \textbf{Zheng-Jun Zha}$^1$\\
$^1$University of Science and Technology of China, Hefei, China\\
$^2$Ant Research, $^3$Alibaba Group, Hangzhou, China\\
$^4$University of California, Berkeley\\
  \texttt{ruilifengustc@gmail.com,}\texttt{\{zkcys001,kevinh\}@mail.ustc.edu.cn,}\\
  \texttt{zhaodeli@gmail.com,}\texttt{jordan@cs.berkeley.edu,}\texttt{zhazj@ustc.edu.cn.} \\
}

\begin{document}
\maketitle

\TOCstop

\begin{abstract}
The rank of neural networks measures information flowing across layers. It is an instance of a key structural condition that applies across broad domains of machine learning. In particular, the assumption of low-rank feature representations leads to algorithmic developments in many architectures. For neural networks, however, the intrinsic mechanism that yields low-rank structures remains vague and unclear. To fill this gap, we perform a rigorous study on the behavior of network rank, focusing particularly on the notion of rank deficiency. We theoretically establish a universal monotonic decreasing property of network rank from the basic rules of differential and algebraic composition, and uncover rank deficiency of network blocks and deep function coupling. By virtue of our numerical tools, we provide the first empirical analysis of the per-layer behavior of network rank in practical settings, \ieno, ResNets, deep MLPs, and Transformers on ImageNet. These empirical results are in direct accord with our theory. Furthermore, we reveal a novel phenomenon of independence deficit caused by the rank deficiency of deep networks, where classification confidence of a given category can be linearly decided by the confidence of a handful of other categories. The theoretical results of this work, together with the empirical findings, may advance understanding of the inherent principles of deep neural networks.
\end{abstract}
\section{Introduction}\label{sec:intro}
In mathematics, the rank of a smooth function measures the volume of independent information captured by the function~\cite{hirsch2012differential}. Deep neural networks are highly smooth functions, thus the rank of a network has long been an essential concept in machine learning that underlies many tasks such as information compression~\cite{vogels2020practical,zha2019rank,mu2020graph,ye2005generalized,wang2005rank}, network pruning~\cite{lin2020hrank,yu2017compressing,blakeney2020pruning,hsu2021language,chen2021drone}, data mining~\cite{cai2013equivalent,hsieh2012low,cheng2014lorslim,zhan2016low,goldfarb2014robust,kheirandishfard2020deep}, computer vision~\cite{zhang2013learning,zhang2013low,kong2017low,jing2015semi,kheirandishfard2020deep,zou2006sparse}, and natural language processing~\cite{chen2018groupreduce,karimi2021compacter,chen2021scatterbrain,chiu2021low}. Numerous methods are either designed to utilize the mathematical property of network ranks, or are derived from an assumption that low-rank structures are to be preferred.

Yet a rigorous investigation to the behavior of rank of general networks, combining both theoretical and empirical arguments, is still absent in current research, weakening our confidence in the being able to predict performance. To the best of our knowledge, there are only a few previous works discussing the rank behavior of specific network architectures, like attention blocks~\cite{ATTnot} and BatchNorms~\cite{BNpreventRC,understandingBN} in pure MLP structures. The empirical validation of those methods are also limited to shallow networks, specific architectures, or merely the final layers of deep networks, leaving the global behavior of general deep neural networks mysterious due to prohibitive space-time complexity for measuring them. 
Rigorous work on network rank that combines both strong theoretical and empirical evidence would have significant implications.

In this paper, we make several contributions towards this challenging goal. We find that the two essential ingredients of deep learning, the chain rules of differential operators and matrix multiplications, are enough to establish a universal principle---that network rank decreases monotonically with the depth of networks. Two factors further enhance the speed of decreasing: a) the explicit rank deficiency of many frequently used network modules, and b) an intrinsic potential of spectrum centralization enforced by the nature of coupling of massive composite functions. To empirically validate our theory, we design numerical tools to efficiently and economically examine the rank behavior of deep neural networks. This is a non-trivial task, as rank is very sensitive to noise and perturbation, and computing ranks of large networks is computationally prohibitive in time and space. Finally, we uncover an interesting phenomenon of independence deficit in multi-class classification networks. We find that many classes do not have their own unique representations in the classification network, and some highly irrelevant classes can decide the outputs of others. This independence deficit can significantly deteriorate the performance of networks in generalized data domains where each class demands a unique representation. In conclusion, the results of this work, together with the numerical tools we invent, may advance understanding of intrinsic properties of deep neural networks, and provide foundations for a broad study of low-dimensional structures in machine learning.
\section{Preliminaries}\label{sec:pre}

\paragraph{Settings}
We consider the general deep neural network with $L$ layers. It is a smooth  vector-valued function $\bm{F}:\mathbb{R}^n\rightarrow\mathbb{R}^d$, where $\mathbb{R}^n$ and $\mathbb{R}^d$ are the ambient space of inputs and outputs, respectively. Deep neural networks are coupling of multiple layers, thus we write $\F$ as:
\begin{equation}
    \F=\bm{f}^L\circ\bm{f}^{L-1}\circ \dots \circ\bm{f}^1.
\end{equation}
For simplicity, we further write the $k$-th sub-network\footnote{In this paper, sub-network means network slice from the input to some intermediate feature layer; layer network means an independent component of the network, without skip connections from the outside to it, like bottleneck layer of ResNet-50.} of $\F$ as
\begin{equation}
    \F_k = \f^k\circ \dots \circ\f^1,
\end{equation}
and we use $\mathcal{F}_k=\F_k(\mathcal{X})$ to denote the feature space of the $k$-th sub-network on the data domain $\mathcal{X}$. We are more interested in the behavior of network rank in the feature spaces rather than scalar outputs (which trivially have rank 1). Thus for classification or regression networks that output a scalar value, we will consider $\F=\F_L$ as the transformation from the input space to the final feature space instead. Thus, we always have $n\gg1$ and $d\gg1$. For example, for ResNet-50~\cite{he2016deep} architecture on ImageNet, we only consider the network slice from the inputs to the last feature layer of 2,048 units. 

\paragraph{Rank of Function}
The rank of a function $\bm{f}=(\f_1, ..., \f_d)^T:\mathbb{R}^n\rightarrow\mathbb{R}^d$ refers to the rank of its Jacobian matrix $J_{\bm{f}}$ over its input domain $\mathcal{X}$, which is defined as 
\begin{equation}
    \rank(\bm{f})=\rank(\bJ_{\bm{f}})=\rank\left((\partial\bm{f}_i (\x) /\partial \x_j)_{n\times d}\right).
\end{equation}
The rank of a function represents the volume of information captured by it in the output~\cite{hirsch2012differential}. That is why it is so important to investigate the behavior of neural networks and many practical applications. Theoretically, by the rank theorem and Sard's theorem of manifolds~\cite{hirsch2012differential}, we can know that rank of the function equals the intrinsic dimension of its output feature space, as captured by the following lemma.\footnote{Due to space limitation,  all the related proofs are attached in the Appendix.}
\begin{lemma}\label{corollary:rank theorem}
Suppose that $\f:\mathbb{R}^n\rightarrow\mathbb{R}^d$ is smooth almost everywhere. Let $\rank(\f)=r$. If data domain $\mathcal{X}$ is a manifold embedded in $\mathbb{R}^n$ and $\bm{\phi}:\mathcal{U}\rightarrow\mathcal{O}$ is a smooth bijective parameterization from an open subset $\mathcal{U}\subset\mathbb{R}^s$ to an open subset $\mathcal{O}\subset\mathcal{X}$, then we have $\dim(\f(\mathcal{X}))=\rank(\bJ_{\f\circ\bm{\phi}})\leq r$. Thus, the rank of function $\f$ gives an upper bound for the intrinsic dimension $\dim(\f(\mathcal{X}))$ of the output space.
\end{lemma}
It is worth mentioning that the intrinsic dimension $\dim(\f(\mathcal{X}))$ of the feature space is usually hard to measure, so the rank of the network gives an operational estimate of it.
\section{Numerical Tools}
Validating the rank behavior of deep neural networks is a challenging task because it involves operations of high complexity on large-scale non-sparse matrices, which is infeasible both in time and space. Computing the full Jacobian representation of sub-networks of ResNet-50, for example, consumes over 150G GPU memory and several days at a single input point. In accuracy, this is even more challenging as rank is very sensitive to small perturbations. The numerical accuracy of $\mathrm{float32}$, $1.19e-7$~\cite{pytorch}, cannot be trivially neglected in computing matrix ranks. Thus, in this section we establish some numerical tools for validating our subsequent arguments, and provide rigorous theoretical support for them. 

\subsection{Numerical Rank: Stable Alternative to Rank}
The rank of large matrices is known to be unstable: it varies significantly under even small noise perturbations~\cite{numericaltextbook}. Matrices perturbed by even small Gaussian noises are almost surely of full rank,
regardless of the true rank of the original matrix. Thus in practice we have to use an alternative: we count the number of  singular values larger than some given threshold $\epsilon$ as the numerical rank of the matrix. Let $\bm{W}\in\mathbb{R}^{n\times d}$ be a given matrix. Its numerical rank with tolerance $\epsilon$ is
\begin{equation}\label{eq:numerical rank}
    \rank_{\epsilon}(\bm{W})=\#\{i\in\mathbb{N}_+: i\leq min\{n,d\}, \sigma_i\geq\epsilon\Vert \bm{W}\Vert_2\},
\end{equation}
where $\Vert \bm{W}\Vert_2$ is the $\ell_2$ norm (spectral norm) of matrix $W$, $\sigma_i,i=1,...,\min\{n,d\}$ are its singular values, and $\#$ is the counting measurement for finite sets. We can prove that the numerical rank is stable under small perturbations. Based on Weyl inequalities~\cite{wely}, we have the following theorem.
\begin{theorem}\label{th:nr_st}
For any given matrix $\bm{W}$,  almost every tolerance $\epsilon>0$, and any perturbation matrix $\bm{D}$, there exists a positive constant $\delta_{\mathrm{max}}(\epsilon)$ such that $\forall\delta\in[0,\delta_{\mathrm{max}}(\epsilon))$, $\rank_{\epsilon}(\bm{W}+\delta\bm{D})=\rank_{\epsilon}(\bm{W})$. If $\bm{W}$ is a low-rank matrix without random perturbations, then there is a $\epsilon_{\mathrm{max}}$ such that for any $\epsilon<\epsilon_{\mathrm{max}}$, $\rank_{\epsilon}(\bm{W}+\delta\bm{D})=\rank_{\epsilon}(\bm{W})=\rank({\bm{W}})$ for all $\delta\in[0,\delta_{\mathrm{max}}(\epsilon))$.
\end{theorem}
This property of the numerical rank metric makes it a suitable tool for investigating the rank behavior of neural networks. Possible small noises can be filtered out in  Jacobian matrices of networks by using numerical rank. It is worth mentioning that random matrices no longer have full rank almost surely under the numerical rank. Instead their rank distribution can be inferred from the well-known Marcenko–Pastur distribution~\cite{marvcenko1967distribution} of random matrices. So under numerical rank, low-rank matrices will be commonly seen. In this paper,  we always use the numerical rank when measuring ranks.

\subsection{Partial Rank of the Jacobian: Estimating Lower Bound of Lost Rank in Deep Networks}\label{sec:partial rank}
To enable the validation of trend of the network ranks, we propose to compute only the rank of sub-matrices of the Jacobian as an alternative. Those sub-matrices are also the Jacobian matrices with respect to a fixed small patch of inputs. Rigorously, given a function $\f$ and its Jacobian $\bJ_{\f}$, we denote partial rank of the Jacobian as the rank of a sub-matrix of the Jacobian that consists of the $j_1$-th, $j_2$-th,...,$j_K$-th column of the original Jacobian
\begin{equation}
    \mathrm{PartialRank}(\bJ_{\f})=\rank(\mathrm{Sub}(\bJ_{\f},j_1,...,j_K))=\rank((\partial\f_i/\partial\x_{j_k})_{d\times K}),
\end{equation}
where $1\leq j_1<\ldots<j_K\leq n.$ We can efficiently compute sub-matrix of the Jacobian by zero padding to small patches of input images. For any data point $\x\in\mathbb{R}^n$, let $\mathrm{Sub}(\x,j_1,...,j_K)=(\x_{j_1},...,\x_{j_K})^T\in\mathbb{R}^K$, and $\bpsi$ pad $\mathrm{Sub}(\x,j_1,...,j_K)$ to the spatial size of $\x$ with zeros: $\bpsi(\mathrm{Sub}(\x,j_1,...,j_K))=(0,..,0,\x_{j_1},0,...,\x_{j_K},0,...,0)^T\in\mathbb{R}^n$ with $\bpsi(\mathrm{Sub}(\x,j_1,...,j_K))_{j_k}=\x_{j_k},k=1,...,K.$ We then have $\bJ_{\f\circ\bpsi}=\mathrm{Sub}(\bJ_{\f},j_1,...,j_K)$. As $K$ can be very small compared with $n$, computing $\bJ_{\f\circ\bpsi}$ can be very cheap in time and space. The partial rank of Jacobian matrices of the network layers measures information captured among the spatial footprint $j_1,...,j_K$ of the original input. They inherit the order relation of the rank of full Jacobian matrices. Thus we can validate the rank diminishing of network Jacobian matrices through the partial rank. 
\begin{lemma}\label{th:partial rank}
For differentiable $\f_1,\f_2$,  $\vert\rank(\f_1)-\rank(\f_2\circ\f_1)\vert\geq \vert\rank(\mathrm{Sub}(\f_1,j_1,\ldots,j_K))-\rank(\mathrm{Sub}(\f_2\circ\f_1,j_1,\ldots,j_K))\vert,\forall 1\leq K\leq n, 1\leq j_1,\dots,j_K\leq n$. Thus variance of partial ranks of adjacent sub-networks gives a lower bound on the variance of their ranks.
\end{lemma}

\subsection{Classification Dimension: Estimating Final Feature Dimension}\label{sec:covdim}
Measuring the intrinsic dimension of feature manifolds is known to be intractable. So we turn to an approximation procedure. For most classification networks, a linear regression over the final feature manifold decides the final network prediction and accuracy. So we can estimate the intrinsic dimension as the minimum number of principal components in the final feature space to preserve a high classification accuracy.
 Given network slice $\F:\mathbb{R}^n\rightarrow\mathbb{R}^d$ from input $\mathcal{X}\subset\mathbb{R}^n$ to final feature space $\F(\mathcal{X})\subset\mathbb{R}^d$, we independently sample $N$ points from random variable $\F(\x),\x\sim\mathbb{P}_{\mathcal{X}}$, where $\mathbb{P}_{\mathcal{X}}$ is the distribution of validation set of data $\mathcal{X}$. We then compute the covariance matrix $\Sigma$ of those $N$ samples, and eigenvectors $\bm{q}^1,\ldots,\bm{q}^d$ of $\Sigma$, sorted by their eigenvalues $\lambda_1\geq\lambda_2\geq\ldots\geq\lambda_d$. Let $\mathrm{cls}:\mathbb{R}^d\rightarrow\mathbb{R}^c$ be the classification predictions based on the final feature representation $\F(\x)$, $\mathrm{Pro}_k$ be projection operator in Euclidean space that projects to the linear subspace spanned by top-$K$ eigenvectors $\bm{q}^1,...,\bm{q}^k,k\leq d$, $\mathbb{P}_{\x,\bm{y}}$ be the joint distribution of sample $\x$ and its label $\bm{y}$, and $\bm{1}_{\mathrm{cond}}$ the indicator for condition $\mathrm{cond}$. The classification dimension is then defined as\vspace{-0.1cm}
\begin{equation}
        \mathrm{ClsDim}(\F(\mathcal{X}))
        =\mathrm{min}_{k}\{k:\mathbb{E}_{(\x,\bm{y})\sim\mathbb{P}_{\mathcal{X},\mathcal{Y}}}[\bm{1}_{\mathrm{Cls}(\mathrm{Pro}_k(\F(\x)))==\bm{y}}]\geq1-\epsilon\},\vspace{-0.06cm}
\end{equation}
which is the minimum dimensionality needed to reconstruct the classification accuracy of the whole model.

\section{Principle of Rank Diminishing}\label{sec:method}
We turn to the \textit{principle of rank diminishing}. We first give a universal justification with minimum limitation on the network, so that we can safely apply this principle to many practical scenarios. 

The principle of rank diminishing describes the behavior of general neural networks with almost everywhere smooth components, which exhibits the monotonic decreasing of network ranks and intrinsic dimensionality of feature manifolds as follows.

\begin{theorem}[Principle of Rank Diminishing]\label{th:principle}
Suppose that each layer $\f_i,i=1,...,L$ of network $\F$ is almost everywhere smooth and  data domain $\mathcal{X}$ is a manifold, then both the rank of sub-networks and intrinsic dimension of feature manifolds decrease monotonically by depth:
\begin{equation}\label{eq:principle1}
    \rank(\f_1)\geq \rank(\f_2\circ\f_1) \geq ...\geq \rank(\f_{L-1}\circ...\circ\f_1)\geq \rank(\F_L),
\end{equation}
\begin{equation}\label{eq:principle2}
    \dim(\mathcal{X})\geq\dim(\mathcal{F}_1)\geq \dim(\mathcal{F}_2)\geq...\geq \dim(\mathcal{F}_L).
\end{equation}
\end{theorem}

\paragraph{Short Argument that the Principle Should Hold Universally.} \cref{th:principle} is ultra intrinsic for deep neural networks. It comes directly from the chain rules of differential operators and basic rules of matrix multiplications. The basic rule of matrix multiplication tells that, for any two matrices $\bA$ and $\bB$, we have $\rank(\bA\bB)\leq\min\{\rank(\bA),\rank(\bB)\}$~\cite{hoffman1971linear}.
Taking this into the chain rule of differential of $\bJ_{\F}=\bJ_{\f^L}\bJ_{\f^{L-1}}...\bJ_{\f^1}$, we then have
    $\rank(\bJ_{\F_k})=\rank(\bJ_{\f^k\circ\F_{k-1}})=\rank(\bJ_{\f^k}\bJ_{\F_{k-1}})\leq\rank(\bJ_{\F_{k-1}}),k=2,...,L,$
which is \cref{eq:principle1}. Applying \cref{corollary:rank theorem} to \cref{eq:principle1} then yields \cref{eq:principle2}.

\begin{figure}[t]
    \centering
    \includegraphics[width=1.0\linewidth]{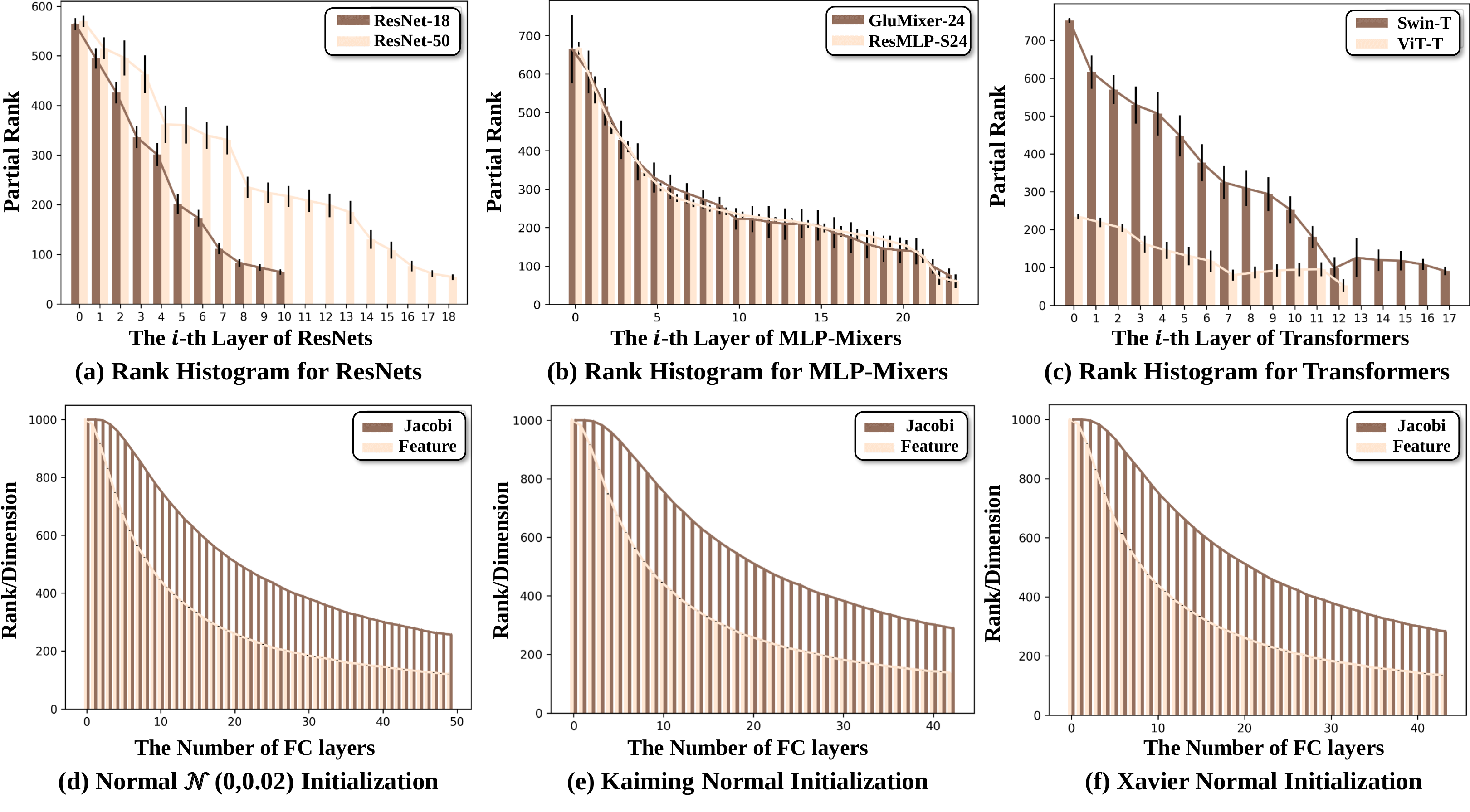}\vspace{-0.3cm}
    \caption{Partial rank of Jacobian matrices of CNN, MLP, and Transformer architecture networks for different layers on ImageNet (top row); rank of Jacobian matrices and feature dimensions of linear MLP network following conditions of \cref{th:lyapunov_gaussian} (bottom rule). All the models show a similar trend of exponential decreasing of ranks as predicted by \cref{th:general_deficiency,th:lyapunov_gaussian}.}
    \label{fig:exp_rank}\vspace{-0.5cm}
\end{figure}

\paragraph{Chance of Equal Sign Holding is Small.} A hypothetical but not practical concern would be that, is it possible that most of the equal signs of \cref{eq:principle1,eq:principle2} hold, so that the rank of network remains no significant dropping throughout the network? This concern can be mitigated by empirical and theoretical arguments. In what follows we will find that, 1) in practice, the rank of sub-networks decreases significantly after applying subsequent layers as shown in \cref{fig:exp_rank}, and 2) in theory, there are two  strong impetuses in deep neural networks to enforce the strict decreasing of ranks which we will discuss in \cref{sec:structural impetus,sec:implicit impetus}.

\subsection{Structural Impetus of Strict Decreasing}\label{sec:structural impetus}
Numerous explicit structures of the network layers can lead to a strict decrease in network ranks. Specifically, the following theorem gives a condition for the strictly greater signs to hold in the principle of rank diminishing.

\begin{table}[t]
    \centering
    \begin{threeparttable}
    \begin{tabular}{ccccccc}
    \toprule
    Arch. & Network & Activ. & \#Param. & Main Block & \#Layer & Top-1 Acc.\\
    \midrule
    \multirow{2}{*}{ResNets} & ResNet-18~\cite{he2016deep} & ReLU~\cite{nair2010rectified} & 11.7M &Bottleneck & 11 & 69.8\%\\
    ~ & ResNet-50~\cite{he2016deep} & ReLU~\cite{nair2010rectified} & 25.6M &Bottleneck& 19 & 76.1\%\\
    \midrule
    \multirow{2}{*}{MLP-like} & GluMixer-24~\cite{shazeer2020glu} & SiLU~\cite{hendrycks2016gaussian} & 25.0M &Mixer-Block& 24 & 78.1\%\\
    ~ & ResMLP-S24~\cite{touvron2021resmlp} & GELU~\cite{hendrycks2016gaussian} & 30.0M &Mixer-Block& 24 & 79.4\%\\
    \midrule
    \multirow{2}{*}{Transformer} & ViT-T~\cite{dosovitskiy2020image} & GELU~\cite{hendrycks2016gaussian} & 5.7M &ViT-Block& 13 & 75.5\%\\
    & Swin-T~\cite{liu2021swin} & GELU~\cite{hendrycks2016gaussian} & 29.0M &Swin-Block& 18 & 81.3\%\\
    \bottomrule
    \end{tabular}
    \end{threeparttable}
    \caption{Information of networks used in empirical validations. All pretrained on ImageNet.}\vspace{-0.4cm}
    \label{tab:network_info}
\end{table}

\begin{figure}[t]
    \centering
   \includegraphics[width=1\linewidth]{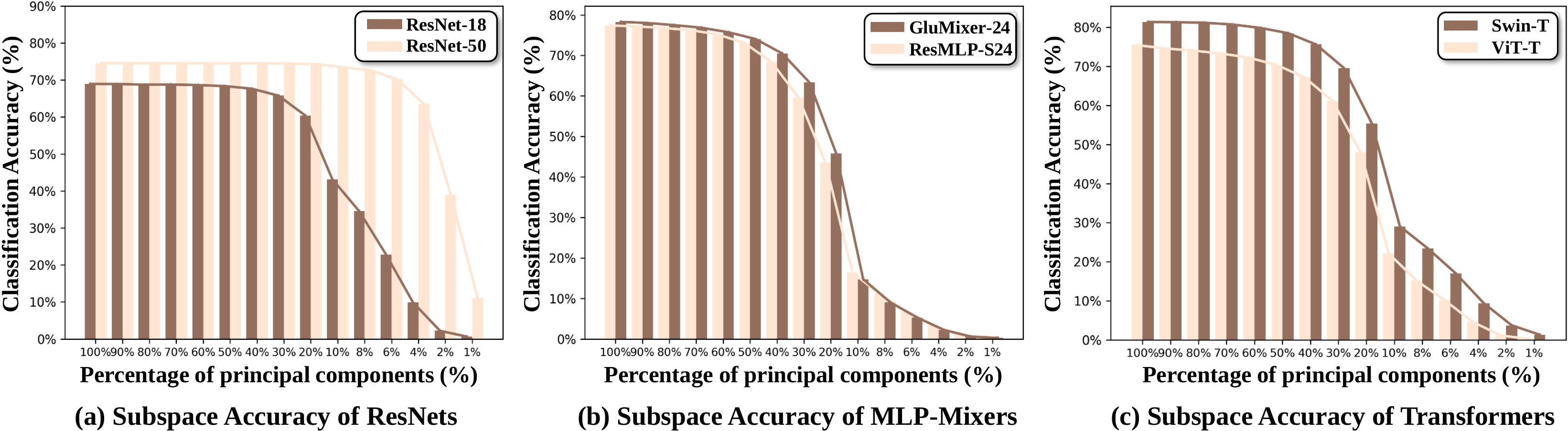}\vspace{-0.25cm}
    \caption{Classification Accuracy (top-1) of using subspaces spanned by top-$k\%$ eigenvectors (principal components) of the final feature manifolds. For all networks a small percentage (see \cref{tab:covdim}) of eigenvectors are enough to reproduce the classification accuracy of the whole network, indicating a low intrinsic dimension of final feature manifolds. \textbf{Note that the $x$-axes are non-linear.}}
    \label{fig:exp_pc_c_three}\vspace{-0.2cm}
\end{figure}

\begin{table}[t]
    \centering
    \begin{tabular}{ccccccc}
    \toprule
        Networks & ResNet-18 & ResNet-50 & GluMixer-24 & ResMLP-S24 & Swin-T & ViT-T\\
        \midrule
        $\mathrm{ClsDim}$ & 149 & 131 & 199 & 196 & 344 & 109\\
        \midrule
        Ambient Dim. & 512 & 2048 & 384 & 384 & 768 & 192\\
    \bottomrule
    \end{tabular}
    \caption{Classification dimensions (with respect to 95\% classification performance of the ambient feature space $\mathbb{R}^d$) and ambient dimensions of the final feature manifolds of different networks. All networks have low intrinsic dimensions for final features. }
    \label{tab:covdim}\vspace{-0.5cm}
\end{table}

\begin{figure}[t]
    \centering
   \includegraphics[width=1\linewidth]{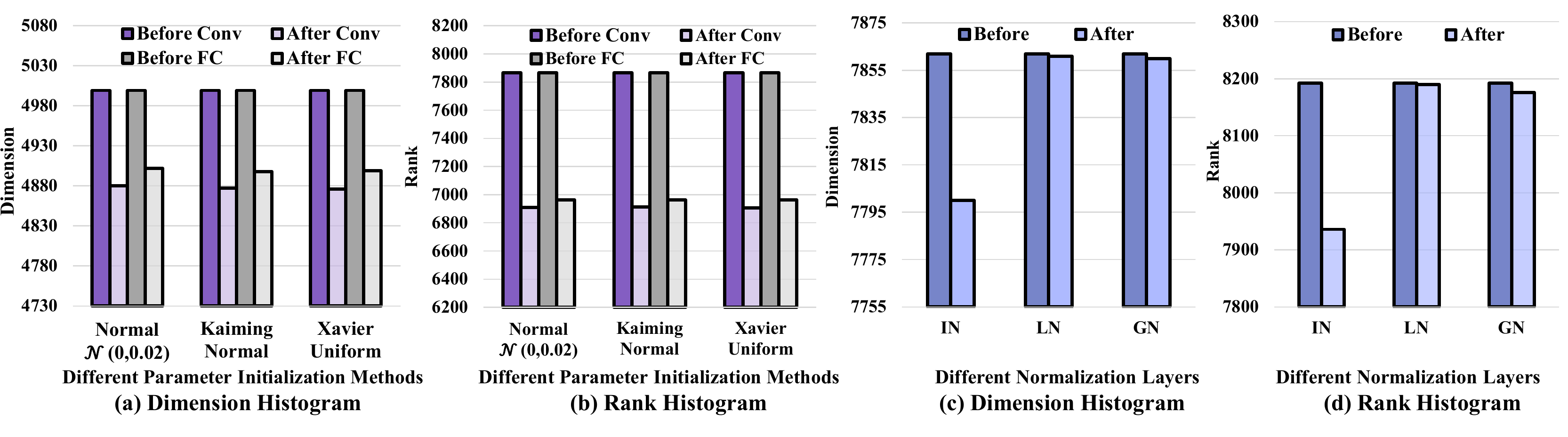}\vspace{-0.3cm}
    \caption{PCA dimension of feature spaces and rank of Jacobian matrix for commonly seen network components under standard Gaussian inputs and randomized weights. Convolution and FC layers tend to lose rank considerably; normalization layers, like InstanceNorm (IN)~\cite{ulyanov2016instance}, LayerNorm (LN)~\cite{ba2016layer}, and GroupNorm (GN)~\cite{wu2018group}, lose rank modestly. But none can preserve rank.}
    \label{fig:exp_structural}\vspace{-0.4cm}
\end{figure}

\begin{theorem}\footnote{The rigorous version is given in the Appendix.}\label{th:structural impetus}
Roughly speaking, if almost everywhere on the input feature manifold, there is a direction such that moving along this direction keeps the output invariant, then the intrinsic dimension of the output feature manifold will be strictly lower than that of the input. The maximum number of independent such directions gives a lower bound on the number of lost intrinsic dimensions.
\end{theorem}
By this theorem, one can immediately find that most frequently used layer designs have high risk in inducing the strict decreasing of network ranks. Normalization layers like LayerNorm~\cite{ba2016layer}, InstanceNorm~\cite{ulyanov2016instance}, and BatchNorm~\cite{ioffe2015batch} may lose dimensions modestly, as the output feature remains invariant along the normalized direction at each point. Linear layers like convolutions, linear transformations (e.g. dense layers), and attentions, can lose rank considerably according to the rank of their weight matrices. They constitute the explicit structural impetus to decrease network ranks and intrinsic dimensions of feature manifolds.

\subsection{Implicit Impetus of Strict Decreasing}\label{sec:implicit impetus}
Apart from the structural impetus we propose in \cref{sec:structural impetus}, there is a more intrinsic strength to pull down network ranks, which we call the implicit impetus. Deep neural networks repeatedly apply layer networks from a fixed function pool (ReLU, MLP, CNN, attention, ResNet block, \etcno.) to the input data and intermediate features to get outputs. Such paradigm accords with the cocycle dynamic systems studied by Lyapunov \textit{et al.}~\cite{temam2012infinite,wolf1985determining}, where the Furstenberg–Kesten theorem \cite{furstenberg1960products} and multiplicative ergodic theorem \cite{pesin1977characteristic} prove that logarithms of singular values divided by evolution time of such chaos system converge to stable constants when time goes to infinity. While products of long chains of matrices are the simplest form of cocycle dynamic systems~\cite{kingman1973subadditive}, we can get an intrinsic impetus of rank collapse tendency of Jacobian matrices independent of network architectures.
\begin{theorem}[Spectrum Centralization of Function Coupling]\label{th:general_deficiency}
Let the network be $\F=\f^L\circ \cdots \circ\f^1$, and all the ambient dimensions of feature manifolds be the same as the ambient dimension of inputs, \ieno, $\f^k:\mathbb{R}^n\rightarrow\mathbb{R}^n, k=1, \ldots,L$. Suppose the Jacobian matrix of each layer $\f_i$ independently follows some distribution $\mu$, and $\mathbb{E}_{\mu}[\max\{\log\Vert\bJ_{\f^k}^{\pm1}\Vert_2,0\}]<\infty$. Let $\sigma_k$ denote the $k$-th largest singular value of $\bJ_{\F}$. Then there is an integer $r<n$ and positive constants $\mu_r,\ldots,\mu_n$ that only depend on $\mu$ such that we have for $\mu$-almost everywhere,
\begin{equation}\label{eq:implicit_impetus}
    \frac{\sigma_k}{\Vert\bJ_{\F}\Vert_2}\sim\exp(-L\mu_k)\rightarrow0,k=r,\ldots,n,~\text{as}~L\rightarrow\infty,
\end{equation}
meaning that for any tolerance $\epsilon>0$, $\rank_{\epsilon}(\F)$ drops below $r+1$ with an exponential speed as $L\rightarrow\infty$.
\end{theorem}

If further assuming that the elements of Jacobian matrices follow Gaussian distributions, we can prove $r=1$ and give a more accurate estimation of constants $\mu_r,...,\mu_n$. As a consequence, we can find that rank of networks collapses to 1 almost surely, which is formalized in the following theorem.

\begin{theorem}\label{th:lyapunov_gaussian}
Let the network be $\F=\f^L\circ...\circ\f^1$, and all the ambient dimensions of feature manifolds be the same as the ambient dimension of inputs, \ieno, $\f^k:\mathbb{R}^n\rightarrow\mathbb{R}^n, k=1,...,L$. Suppose that $\bJ_{\f^i}$ independently follows the standard Gaussian distribution. Let $\sigma_k$ denote the $k$-th largest singular value of $\bJ_{\F}$. Then almost surely
\begin{equation}
    \lim_{L\rightarrow\infty}\left(\frac{\sigma_k}{\Vert\bJ_{\F}\Vert_2}\right)^{\frac{1}{L}}=\exp{\frac{1}{2}\left(\psi(\frac{n-k+1}{2})-\psi(\frac{n}{2})\right)}<1, k=2,\ldots,n,
\end{equation}
where $\psi=\Gamma/\Gamma'$ and $\Gamma$ is the Gamma function. That means for a large $L$ and any tolerance $\epsilon$, $\rank_{\epsilon}(\F)$ drops to $1$ exponentially with speed $nC^{L}$, where $C<1$ is a positive constant that only depends on $n$.
\end{theorem}

\paragraph{Connection with Gradient Explosion} Bengio \textit{et. al.} \cite{bengio1994learning,pascanu2013difficulty} discuss the gradient explosion issue of deep neural networks, where the largest singular value of the Jacobian matrix tends to infinity when the layer gets deeper. This problem  could be viewed as a special case of \cref{th:lyapunov_gaussian} that investigates the behavior of all singular values of  deep neural networks. The behavior of network ranks in fact manipulates the well-known gradient explosion issue. Rigorously, we have the following conclusion.
\begin{corollary}\label{corollary:ge}
Under the condition of \cref{th:lyapunov_gaussian}, then almost surely gradient explosion happens at an exponential speed, \ieno,  $\log\Vert\bJ_{\F}\Vert_2=\log\sigma_1\sim \frac{L}{2}(\log2+\psi(n/2))\rightarrow\infty$ when $L$ is large.
\end{corollary}

\subsection{Validation}
\label{Validations}
\paragraph{Setup} In this section, we validate our theory in three types of architectures of benchmark deep neural networks, CNNs, MLPs, and Transformers, in the ImageNet~\cite{deng2009imagenet} data domain. Information of those networks is listed in \cref{tab:network_info}. For validating the tendency of network rank of Jacobian matrices, we use the numerical rank of sub-matrices of Jacobian on the central $16\times16\times3$ image patch of input images. We report the results of other choices of patches in the Appendix. When measuring rank, we set $\epsilon=\mathrm{eps}\times N$, where $\mathrm{eps}$ is the digital accuracy of $\mathrm{float32}$ (\ieno, $1.19e-7$) and $N$ is the number of singular values of the matrix to measure. This threshold represents the minimum digital accuracy of numerical rank we can capture in data stored as $\mathrm{float32}$. All the experiments are conducted on the validation set of ImageNet and NVIDIA A100-SXM-80G GPUs.

\paragraph{Rank Diminishing of Jacobians} As is discussed in \cref{sec:partial rank,th:partial rank}, the partial rank of the Jacobian is a powerful weapon for us to detect the behavior of huge Jacobian matrices, which are infeasible to compute in practice. The decent value of partial ranks of adjacent sub-networks provides a lower bound to that of full ranks of them. \cref{fig:exp_rank} (a,b,c) report the partial rank of Jacobian matrices of three types of architectures, where we can find consistent diminishing of partial ranks in each layer, indicating  a larger rank losing for the full rank of Jacobian matrices. 

\paragraph{Intrinsic Dimension of the Final Feature Manifold}
To get a further estimation of how many dimensions remain in the final feature representation, we measure the classification dimension in \cref{fig:exp_pc_c_three,tab:covdim}. We report the classification accuracy produced by projecting final feature representations to its top $k\%$ eigenvectors in \cref{fig:exp_pc_c_three}. We choose a threshold of $\epsilon$ such that this procedure can reproduce $95\%$ of the original accuracy of the network. The corresponding $\mathrm{ClsDim}$ is reported in \cref{tab:covdim}. As discussed in \cref{sec:covdim}, this gives an estimation of the intrinsic dimension of the final feature manifold. We can find a universal low-rank structure for all types of networks.

\paragraph{Implicit Impetus} \cref{th:lyapunov_gaussian} gives an exponential speed of rank decent by layers. We find that it corresponds well with practice. We investigate this exponential law in a toy network of MLP-50, which is composed of 50 dense layers, each with 1,000 hidden units. The MLP-50 network takes Gaussian noise vectors of $\mathbb{R}^{1000}$ as inputs, and returns a prediction of 1,000 categories. As all the feature manifolds are linear subspaces in this case, their intrinsic dimensions can be directly measured by the numerical rank of their covariance matrices. We report the full rank of Jacobian matrices and intrinsic dimensions of feature manifolds under three different randomly chosen weights in \cref{fig:exp_rank} (d,e,f). Due to the digital accuracy of $\mathrm{float32}$, we stop calculation in each setting when the absolute values of elements of the matrices are lower than $1.19e-7$. We can find  standard curves of exponential laws in all cases for both ranks of Jacobian and intrinsic dimensions of features. By comparison, we can further find that the ranks of benchmark deep neural networks on ImageNet bear a striking resemblance to the exponential laws of our toy setting, which confirms the proposed implicit impetus in those models.

\paragraph{Structural Impetus} We validate the structural impetus in~\cref{fig:exp_structural}. To give an estimation for general cases, here we use Gaussian noises with the size of $128 \times 8 \times 8$ as inputs, and randomize weights of the network components to be validated. We plug those components into a simple fully-connected (FC) layer of 8,192 hidden units. As the structure is simple, we directly measure the intrinsic dimension of feature spaces and the full rank of Jacobian matrices before and after the features pass the network components to be measured. The dimension is determined by the number of PCA eigenvalues \cite{hotelling1936relations,wold1987principal} larger than $1.19e-7\times N\times \sigma_{\mathrm{max}}$, where $N$ is the number of PCA eigenvalues, and $\sigma_{\mathrm{max}}$ is the largest PCA eigenvalue. The batch size is set to 5,000. We find that the convolution (the kernel size is $3 \times 3$) and FC layers (the weight size is 8,192) tend to lose rank considerably, while different normalization layers also lose rank modestly. But none of them can preserve rank invariant.

\paragraph{Possible Remission Approaches to Rank Diminishing} 
There are quite some techniques, at least in theory, can remiss the network rank diminishing. Typical examples are skip connection~\cite{ATTnot} and BatchNorm~\cite{BNpreventRC}, which we will discuss in the Appendix due to page limitation.


\section{Independence Deficit of Final Feature Manifolds}\label{sec:discussion}
\definecolor{cf}{RGB}{53, 136, 195}
\definecolor{or}{RGB}{84, 130, 53}

\begin{figure}[t]
\centering
\includegraphics[width=1\linewidth]{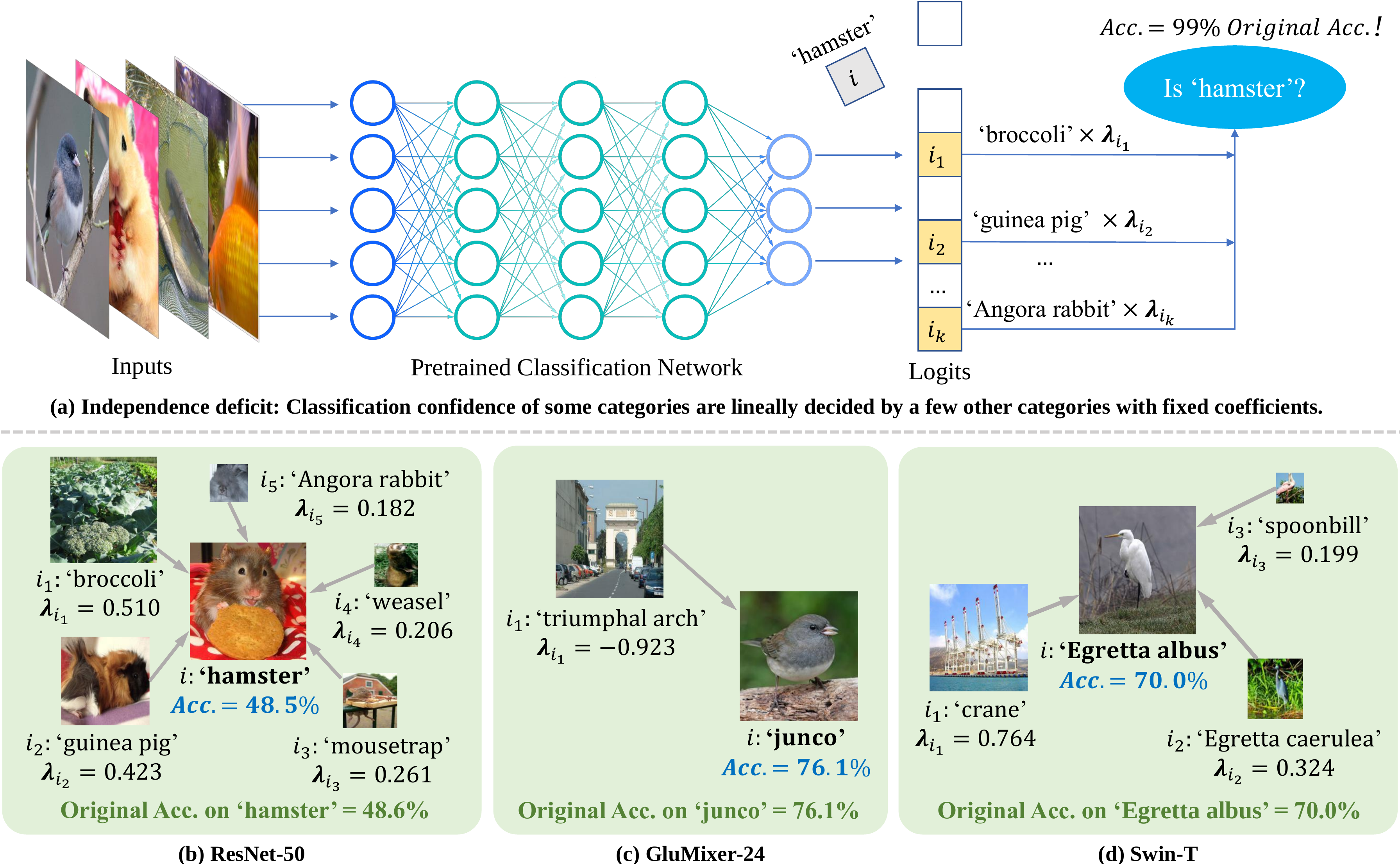}\vspace{-0.3cm}
\caption{Independence deficit. Classification confidence of some ImageNet categories are lineally decided by a few other categories with fixed coefficients in the whole data domain. We illustrate this phenomenon in (a). Here we present some results from ResNet-50, GluMixer-24, and Swin-T. In the (b,c,d) we illustrate the categories of $i_1,...,i_k$ (in the surrounding) to linearly decide category $i$ (in the center) and their corresponding weights $\lambda_{i_1},...,\lambda_{i_k}$. The classification accuracy on the validation set of using \cref{eq:hamster}, instead of the true logits, to predict the label is reported in \textbf{\textcolor{cf}{blue}} (if tested on positive samples only, the accuracy rates are 98\%, 90\%, 82\% for cases in (b,c,d) correspondingly). For comparison, the original accuracy for the corresponding categories are reported in \textbf{\textcolor{or}{green}}. We can find that 1) a few other categories can decide the confidence of the target category $i$; 2) some very irrelevant categories contribute the largest weights. For example in (c), the logits of class `junco' is the negative of `triumphal arch'. Both of them indicate a rather drastic competition of different categories for independent representations in final features due to the tight rank budgets.}
\label{fig:hamster}\vspace{-0.5cm}
\end{figure}

In this section, we provide a further perspective to study the low-rank structure of the final feature manifold, which induces an interesting finding of independence deficit in deep neural networks. We have already known that the final feature representations of deep neural networks admit a very low intrinsic dimension. Thus there are only a few independent representations to decide the classification scores for all the 1,000 categories of ImageNet. It is then curious whether we can predict the outputs of the network for some categories based on the outputs for a few other categories, as  illustrated in \cref{fig:hamster} (a). And if we can, will those categories be strongly connected to each other? A surprising fact is that, we can find many counter examples of irrelevant categories dominating the network outputs for given categories regarding various network architectures. This interesting phenomenon indicates a rather drastic competing in the final feature layer for the tight rank budgets of all categories, which yields non-realistic dependencies of different categories. 

To find the dependencies of categories in final features, we can solve the following Lasso problem~\cite{tibshirani1996regression},
\begin{equation}
    \bm{\lambda}^*=\mathop{\arg\min}_{\bm{\lambda}_i=-1}\mathbb{E}_{\x}[\Vert\bm{\lambda}^T \bm{W}\F(\x)\Vert_2^2]+\eta\Vert\bm{\lambda}\Vert_1,
\end{equation}
where $\F(\x)\in\mathbb{R}^{1000}$ is the slice of network from inputs to the final feature representation, $\x$ is the sample from ImageNet $\mathcal{X}$, and $\bm{W}$ is the final dense layer. The solution $\bm{\lambda}^*$ will be a sparse vector, with $k$ non-zero elements $\lambda_{i_1}\geq\lambda_{i_2}\geq...\geq\lambda_{i_k},k\ll1000$. We can then get 
\begin{equation}\label{eq:hamster}
    \mathrm{logits}(\x, i)\approx\bm{\lambda}_{i_1}\mathrm{logits}(\x,i_1)+...+\bm{\lambda}_{i_k}\mathrm{logits}(\x,i_k), i\notin\{i_1,...,i_k\},k\ll1000,\forall\x\in\mathcal{X},
\end{equation}
where $\mathrm{logits}(\x,i_j),j=1,\ldots,k$ is the logits of network for category $i_j$, \ieno, $\mathrm{logits}(\x,i_j)=\bm{W}_{i_j}\F(\x)$. It is easy to see that outputs for category $i$ are linearly decided by outputs for $i_1,...,i_k$ and are dominated by outputs for $i_1$.

In \cref{fig:hamster} we demonstrate the solutions of \cref{eq:hamster} for three different categories in ImageNet with $\eta=20$, and network architectures ResNet-50, GluMixer-24, and Swin-T. The results are surprising. It shows that many categories of the network predictions are in fact `redundant', as they are purely decided by the predictions of the other categories with simple linear coefficients. In this case, the entanglement of different categories cannot be avoided, thus the network may perform poorly under domain shift. An even more surprising finding is that, some  very irrelevant categories hold the largest weights when deciding the predictions of the redundant categories, which means that the networks just neglect the unique representations of those categories in training and yield  over-fitting when predicting them.

\section{Related Work}
Previous studies of rank deficiency in deep neural networks follow two parallel clues. One is the study of rank behavior in specific neural network architectures. \cite{ATTnot} studies  deep networks consisting of pure self-attention networks, and proves that they converge exponentially to a rank-1 matrix under the assumption of globally bounded  weight matrices. \cite{BNpreventRC} studies the effect of BatchNorm on MLPs and shows that BatchNorm can prevent drastic diminishing of network ranks in some small networks and datasets. Both of those works avoid directly validating the behavior of network ranks in intermediate layers due to the lacking of efficient numerical tools. An independent clue is the study of implicit self-regularization, which finds that weight matrices tend to lose ranks after training. \cite{martin2021implicit} studies this phenomenon in infinitely-wide, over-parametric neural networks with tools from random matrix theory. \cite{arora2019implicit} studies this phenomenon in deep matrix decomposition. Those works focus on the theoretical behavior of rank of weight matrices induced by the training instead of network ranks.
\section{Conclusion}\label{sec:conclusion}
This paper studies the rank behavior of deep neural networks. In contrast to  previous work, we focus on directly validating rank behavior with deep neural networks of diverse benchmarks and various settings for real scenarios. We first formalize the analysis and measurement of network ranks.
Then under the proposed numerical tools and theoretical analysis, we demonstrate the universal rank diminishing of deep neural networks from both empirical and theoretical perspectives. We further support the rank-deficient structure of networks by revealing the independence deficit phenomenon, where network predictions for a category can be linearly decided by a few other, even irrelevant categories. The results of this work may advance understanding of the behavior of fundamental network architectures and provide intuition for a wide range of work pertaining to network ranks.
{\small
\bibliographystyle{abbrvnat}
\bibliography{ref}

\begin{thebibliography}{60}
\providecommand{\natexlab}[1]{#1}
\providecommand{\url}[1]{\texttt{#1}}
\expandafter\ifx\csname urlstyle\endcsname\relax
  \providecommand{\doi}[1]{doi: #1}\else
  \providecommand{\doi}{doi: \begingroup \urlstyle{rm}\Url}\fi

\bibitem[Arora et~al.(2019)Arora, Cohen, Hu, and Luo]{arora2019implicit}
S.~Arora, N.~Cohen, W.~Hu, and Y.~Luo.
\newblock Implicit regularization in deep matrix factorization.
\newblock \emph{Adv. Neural Inform. Process. Syst.}, 32, 2019.

\bibitem[Ba et~al.(2016)Ba, Kiros, and Hinton]{ba2016layer}
J.~L. Ba, J.~R. Kiros, and G.~E. Hinton.
\newblock Layer normalization.
\newblock \emph{arXiv preprint arXiv:1607.06450}, 2016.

\bibitem[Bengio et~al.(1994)Bengio, Simard, and Frasconi]{bengio1994learning}
Y.~Bengio, P.~Simard, and P.~Frasconi.
\newblock Learning long-term dependencies with gradient descent is difficult.
\newblock \emph{IEEE Trans. Neural Netw.}, 5\penalty0 (2):\penalty0 157--166,
  1994.

\bibitem[Bjorck et~al.(2018)Bjorck, Gomes, Selman, and
  Weinberger]{understandingBN}
N.~Bjorck, C.~P. Gomes, B.~Selman, and K.~Q. Weinberger.
\newblock Understanding batch normalization.
\newblock \emph{Adv. Neural Inform. Process. Syst.}, 31, 2018.

\bibitem[Blakeney et~al.(2020)Blakeney, Yan, and Zong]{blakeney2020pruning}
C.~Blakeney, Y.~Yan, and Z.~Zong.
\newblock Is pruning compression?: Investigating pruning via network layer
  similarity.
\newblock In \emph{IEEE Winter Conf. Appl. Comput. Vis.}, pages 914--922, 2020.

\bibitem[Cai et~al.(2013)Cai, Ding, Nie, and Huang]{cai2013equivalent}
X.~Cai, C.~Ding, F.~Nie, and H.~Huang.
\newblock On the equivalent of low-rank linear regressions and linear
  discriminant analysis based regressions.
\newblock In \emph{Proc. ACM SIGKDD Int. Conf. Knowl. Discov. Data Min.}, pages
  1124--1132, 2013.

\bibitem[Chen et~al.(2021{\natexlab{a}})Chen, Dao, Winsor, Song, Rudra, and
  R{\'e}]{chen2021scatterbrain}
B.~Chen, T.~Dao, E.~Winsor, Z.~Song, A.~Rudra, and C.~R{\'e}.
\newblock Scatterbrain: Unifying sparse and low-rank attention.
\newblock \emph{Adv. Neural Inform. Process. Syst.}, 34, 2021{\natexlab{a}}.

\bibitem[Chen et~al.(2018)Chen, Si, Li, Chelba, and Hsieh]{chen2018groupreduce}
P.~Chen, S.~Si, Y.~Li, C.~Chelba, and C.-J. Hsieh.
\newblock Groupreduce: Block-wise low-rank approximation for neural language
  model shrinking.
\newblock \emph{Adv. Neural Inform. Process. Syst.}, 31, 2018.

\bibitem[Chen et~al.(2021{\natexlab{b}})Chen, Yu, Dhillon, and
  Hsieh]{chen2021drone}
P.~Chen, H.-F. Yu, I.~Dhillon, and C.-J. Hsieh.
\newblock Drone: Data-aware low-rank compression for large {NLP} models.
\newblock \emph{Adv. Neural Inform. Process. Syst.}, 34:\penalty0 29321--29334,
  2021{\natexlab{b}}.

\bibitem[Cheng et~al.(2014)Cheng, Yin, and Yu]{cheng2014lorslim}
Y.~Cheng, L.~Yin, and Y.~Yu.
\newblock Lorslim: Low rank sparse linear methods for top-n recommendations.
\newblock In \emph{IEEE Int. Conf. on Data Min.}, pages 90--99. IEEE, 2014.

\bibitem[Chiu et~al.(2021)Chiu, Deng, and Rush]{chiu2021low}
J.~Chiu, Y.~Deng, and A.~Rush.
\newblock Low-rank constraints for fast inference in structured models.
\newblock \emph{Adv. Neural Inform. Process. Syst.}, 34, 2021.

\bibitem[Daneshmand et~al.(2020)Daneshmand, Kohler, Bach, Hofmann, and
  Lucchi]{BNpreventRC}
H.~Daneshmand, J.~Kohler, F.~Bach, T.~Hofmann, and A.~Lucchi.
\newblock Batch normalization provably avoids ranks collapse for randomly
  initialised deep networks.
\newblock \emph{Adv. Neural Inform. Process. Syst.}, 33:\penalty0 18387--18398,
  2020.

\bibitem[Deng et~al.(2009)Deng, Dong, Socher, Li, Li, and
  Fei-Fei]{deng2009imagenet}
J.~Deng, W.~Dong, R.~Socher, L.-J. Li, K.~Li, and L.~Fei-Fei.
\newblock Image{N}et: A large-scale hierarchical image database.
\newblock In \emph{IEEE Conf. Comput. Vis. Pattern Recog.}, pages 248--255.
  Ieee, 2009.

\bibitem[Dong et~al.(2021)Dong, Cordonnier, and Loukas]{ATTnot}
Y.~Dong, J.-B. Cordonnier, and A.~Loukas.
\newblock Attention is not all you need: Pure attention loses rank doubly
  exponentially with depth.
\newblock In \emph{Int. Conf. Mach. Learn.}, pages 2793--2803. PMLR, 2021.

\bibitem[Dosovitskiy et~al.(2020)Dosovitskiy, Beyer, Kolesnikov, Weissenborn,
  Zhai, Unterthiner, Dehghani, Minderer, Heigold, Gelly,
  et~al.]{dosovitskiy2020image}
A.~Dosovitskiy, L.~Beyer, A.~Kolesnikov, D.~Weissenborn, X.~Zhai,
  T.~Unterthiner, M.~Dehghani, M.~Minderer, G.~Heigold, S.~Gelly, et~al.
\newblock An image is worth 16x16 words: Transformers for image recognition at
  scale.
\newblock In \emph{Int. Conf. Learn. Represent.}, 2020.

\bibitem[Furstenberg and Kesten(1960)]{furstenberg1960products}
H.~Furstenberg and H.~Kesten.
\newblock Products of random matrices.
\newblock \emph{Ann. Math. Stat.}, 31\penalty0 (2):\penalty0 457--469, 1960.

\bibitem[Furstenberg and Kesten(1983)]{FKT}
H.~Furstenberg and H.~Kesten.
\newblock Random matrix products and measures on projective spaces.
\newblock \emph{Israel J. Math.}, 46:\penalty0 12--32, 1983.

\bibitem[Goldfarb and Qin(2014)]{goldfarb2014robust}
D.~Goldfarb and Z.~Qin.
\newblock Robust low-rank tensor recovery: Models and algorithms.
\newblock \emph{SIAM J. Matrix Anal. Appl.}, 35\penalty0 (1):\penalty0
  225--253, 2014.

\bibitem[He et~al.(2016)He, Zhang, Ren, and Sun]{he2016deep}
K.~He, X.~Zhang, S.~Ren, and J.~Sun.
\newblock Deep residual learning for image recognition.
\newblock In \emph{IEEE Conf. Comput. Vis. Pattern Recog.}, pages 770--778,
  2016.

\bibitem[Hendrycks and Gimpel(2016)]{hendrycks2016gaussian}
D.~Hendrycks and K.~Gimpel.
\newblock Gaussian error linear units ({GELU}s).
\newblock \emph{arXiv preprint arXiv:1606.08415}, 2016.

\bibitem[Hirsch(2012)]{hirsch2012differential}
M.~W. Hirsch.
\newblock \emph{Differential topology}, volume~33.
\newblock Springer Science \& Business Media, 2012.

\bibitem[Hoffman(1971)]{hoffman1971linear}
K.~Hoffman.
\newblock \emph{Linear algebra}.
\newblock Englewood Cliffs, NJ, Prentice-Hall, 1971.

\bibitem[Hotelling(1936)]{hotelling1936relations}
H.~Hotelling.
\newblock Relations between two sets of variates.
\newblock \emph{Biometrika}, 28\penalty0 (3/4):\penalty0 321--377, 1936.

\bibitem[Hsieh et~al.(2012)Hsieh, Chiang, and Dhillon]{hsieh2012low}
C.-J. Hsieh, K.-Y. Chiang, and I.~S. Dhillon.
\newblock Low rank modeling of signed networks.
\newblock In \emph{Proc. ACM SIGKDD Int. Conf. Knowl. Discov. Data Min.}, pages
  507--515, 2012.

\bibitem[Hsu et~al.(2021)Hsu, Hua, Chang, Lou, Shen, and Jin]{hsu2021language}
Y.-C. Hsu, T.~Hua, S.~Chang, Q.~Lou, Y.~Shen, and H.~Jin.
\newblock Language model compression with weighted low-rank factorization.
\newblock In \emph{Int. Conf. Learn. Represent.}, 2021.

\bibitem[Ioffe and Szegedy(2015)]{ioffe2015batch}
S.~Ioffe and C.~Szegedy.
\newblock Batch normalization: Accelerating deep network training by reducing
  internal covariate shift.
\newblock In \emph{Int. Conf. Mach. Learn.}, pages 448--456. PMLR, 2015.

\bibitem[Jing et~al.(2015)Jing, Yang, Yu, and Ng]{jing2015semi}
L.~Jing, L.~Yang, J.~Yu, and M.~K. Ng.
\newblock Semi-supervised low-rank mapping learning for multi-label
  classification.
\newblock In \emph{IEEE Conf. Comput. Vis. Pattern Recog.}, pages 1483--1491,
  2015.

\bibitem[Karimi~Mahabadi et~al.(2021)Karimi~Mahabadi, Henderson, and
  Ruder]{karimi2021compacter}
R.~Karimi~Mahabadi, J.~Henderson, and S.~Ruder.
\newblock Compacter: Efficient low-rank hypercomplex adapter layers.
\newblock \emph{Adv. Neural Inform. Process. Syst.}, 34, 2021.

\bibitem[Kheirandishfard et~al.(2020)Kheirandishfard, Zohrizadeh, and
  Kamangar]{kheirandishfard2020deep}
M.~Kheirandishfard, F.~Zohrizadeh, and F.~Kamangar.
\newblock Deep low-rank subspace clustering.
\newblock In \emph{IEEE Conf. Comput. Vis. Pattern Recog. Worksh.}, pages
  864--865, 2020.

\bibitem[Kingman(1973)]{kingman1973subadditive}
J.~F.~C. Kingman.
\newblock Subadditive ergodic theory.
\newblock \emph{Ann. Probab.}, pages 883--899, 1973.

\bibitem[Kong and Fowlkes(2017)]{kong2017low}
S.~Kong and C.~Fowlkes.
\newblock Low-rank bilinear pooling for fine-grained classification.
\newblock In \emph{IEEE Conf. Comput. Vis. Pattern Recog.}, pages 365--374,
  2017.

\bibitem[Lin et~al.(2020)Lin, Ji, Wang, Zhang, Zhang, Tian, and
  Shao]{lin2020hrank}
M.~Lin, R.~Ji, Y.~Wang, Y.~Zhang, B.~Zhang, Y.~Tian, and L.~Shao.
\newblock Hrank: Filter pruning using high-rank feature map.
\newblock In \emph{IEEE Conf. Comput. Vis. Pattern Recog.}, pages 1529--1538,
  2020.

\bibitem[Liu et~al.(2021)Liu, Lin, Cao, Hu, Wei, Zhang, Lin, and
  Guo]{liu2021swin}
Z.~Liu, Y.~Lin, Y.~Cao, H.~Hu, Y.~Wei, Z.~Zhang, S.~Lin, and B.~Guo.
\newblock Swin transformer: Hierarchical vision transformer using shifted
  windows.
\newblock In \emph{IEEE Conf. Comput. Vis. Pattern Recog.}, pages 10012--10022,
  2021.

\bibitem[Mar{\v{c}}enko and Pastur(1967)]{marvcenko1967distribution}
V.~A. Mar{\v{c}}enko and L.~A. Pastur.
\newblock Distribution of eigenvalues for some sets of random matrices.
\newblock \emph{Math. USSR Sb.}, 1\penalty0 (4):\penalty0 457, 1967.

\bibitem[Martin and Mahoney(2021)]{martin2021implicit}
C.~H. Martin and M.~W. Mahoney.
\newblock Implicit self-regularization in deep neural networks: Evidence from
  random matrix theory and implications for learning.
\newblock \emph{J. Mach. Learn. Res.}, 22\penalty0 (165):\penalty0 1--73, 2021.

\bibitem[Mu et~al.(2020)Mu, Xiong, Fan, Liu, Wu, and Gao]{mu2020graph}
J.~Mu, R.~Xiong, X.~Fan, D.~Liu, F.~Wu, and W.~Gao.
\newblock Graph-based non-convex low-rank regularization for image compression
  artifact reduction.
\newblock \emph{IEEE Trans. Image Process.}, 29:\penalty0 5374--5385, 2020.

\bibitem[Nair and Hinton(2010)]{nair2010rectified}
V.~Nair and G.~E. Hinton.
\newblock Rectified linear units improve restricted {B}oltzmann machines.
\newblock In \emph{Int. Conf. Mach. Learn.}, pages 807--814. PMLR, 2010.

\bibitem[Newman(1986)]{newman1986distribution}
C.~M. Newman.
\newblock The distribution of {L}yapunov exponents: Exact results for random
  matrices.
\newblock \emph{Commun. Math. Phys.}, 103\penalty0 (1):\penalty0 121--126,
  1986.

\bibitem[Pascanu et~al.(2013)Pascanu, Mikolov, and
  Bengio]{pascanu2013difficulty}
R.~Pascanu, T.~Mikolov, and Y.~Bengio.
\newblock On the difficulty of training recurrent neural networks.
\newblock In \emph{Int. Conf. Mach. Learn.}, pages 1310--1318. PMLR, 2013.

\bibitem[Paszke et~al.(2019)Paszke, Gross, Massa, Lerer, Bradbury, Chanan,
  Killeen, Lin, Gimelshein, Antiga, et~al.]{pytorch}
A.~Paszke, S.~Gross, F.~Massa, A.~Lerer, J.~Bradbury, G.~Chanan, T.~Killeen,
  Z.~Lin, N.~Gimelshein, L.~Antiga, et~al.
\newblock Pytorch: An imperative style, high-performance deep learning library.
\newblock \emph{Adv. Neural Inform. Process. Syst.}, 32, 2019.

\bibitem[Pesin(1977)]{pesin1977characteristic}
Y.~B. Pesin.
\newblock Characteristic {L}yapunov exponents and smooth ergodic theory.
\newblock \emph{Russ. Math. Surv.}, 32\penalty0 (4):\penalty0 55, 1977.

\bibitem[Quarteroni et~al.(2010)Quarteroni, Sacco, and
  Saleri]{numericaltextbook}
A.~Quarteroni, R.~Sacco, and F.~Saleri.
\newblock \emph{Numerical mathematics}, volume~37.
\newblock Springer Science \& Business Media, 2010.

\bibitem[Shazeer(2020)]{shazeer2020glu}
N.~Shazeer.
\newblock G{LU} variants improve transformer.
\newblock \emph{arXiv preprint arXiv:2002.05202}, 2020.

\bibitem[Temam(2012)]{temam2012infinite}
R.~Temam.
\newblock \emph{Infinite-dimensional dynamical systems in mechanics and
  physics}, volume~68.
\newblock Springer Science \& Business Media, 2012.

\bibitem[Tibshirani(1996)]{tibshirani1996regression}
R.~Tibshirani.
\newblock Regression shrinkage and selection via the lasso.
\newblock \emph{J. R. Stat. Soc. Series B Stat. Methodol.}, 58\penalty0
  (1):\penalty0 267--288, 1996.

\bibitem[Touvron et~al.(2021)Touvron, Bojanowski, Caron, Cord, El-Nouby, Grave,
  Izacard, Joulin, Synnaeve, Verbeek, et~al.]{touvron2021resmlp}
H.~Touvron, P.~Bojanowski, M.~Caron, M.~Cord, A.~El-Nouby, E.~Grave,
  G.~Izacard, A.~Joulin, G.~Synnaeve, J.~Verbeek, et~al.
\newblock Res{MLP}: Feedforward networks for image classification with
  data-efficient training.
\newblock \emph{arXiv preprint arXiv:2105.03404}, 2021.

\bibitem[Ulyanov et~al.(2016)Ulyanov, Vedaldi, and
  Lempitsky]{ulyanov2016instance}
D.~Ulyanov, A.~Vedaldi, and V.~Lempitsky.
\newblock Instance normalization: The missing ingredient for fast stylization.
\newblock \emph{arXiv preprint arXiv:1607.08022}, 2016.

\bibitem[Vogels et~al.(2020)Vogels, Karimireddy, and
  Jaggi]{vogels2020practical}
T.~Vogels, S.~P. Karimireddy, and M.~Jaggi.
\newblock Practical low-rank communication compression in decentralized deep
  learning.
\newblock \emph{Adv. Neural Inform. Process. Syst.}, 33:\penalty0 14171--14181,
  2020.

\bibitem[Wang and Ahuja(2005)]{wang2005rank}
H.~Wang and N.~Ahuja.
\newblock Rank-r approximation of tensors using image-as-matrix representation.
\newblock In \emph{IEEE Conf. Comput. Vis. Pattern Recog.}, volume~2, pages
  346--353. IEEE, 2005.

\bibitem[Weyl(1912)]{wely}
H.~Weyl.
\newblock Das asymptotische verteilungsgesetz der eigenwerte linearer
  partieller differentialgleichungen.
\newblock \emph{Math. Ann.}, 71:\penalty0 441--479, 1912.

\bibitem[Wold et~al.(1987)Wold, Esbensen, and Geladi]{wold1987principal}
S.~Wold, K.~Esbensen, and P.~Geladi.
\newblock Principal component analysis.
\newblock \emph{Chemom. Intell. Lab. Syst}, 2\penalty0 (1-3):\penalty0 37--52,
  1987.

\bibitem[Wolf et~al.(1985)Wolf, Swift, Swinney, and
  Vastano]{wolf1985determining}
A.~Wolf, J.~B. Swift, H.~L. Swinney, and J.~A. Vastano.
\newblock Determining {L}yapunov exponents from a time series.
\newblock \emph{Physica D}, 16\penalty0 (3):\penalty0 285--317, 1985.

\bibitem[Wu and He(2018)]{wu2018group}
Y.~Wu and K.~He.
\newblock Group normalization.
\newblock In \emph{Eur. Conf. Comput. Vis.}, pages 3--19, 2018.

\bibitem[Ye(2005)]{ye2005generalized}
J.~Ye.
\newblock Generalized low rank approximations of matrices.
\newblock \emph{Mach. Learn.}, 61\penalty0 (1):\penalty0 167--191, 2005.

\bibitem[Yu et~al.(2017)Yu, Liu, Wang, and Tao]{yu2017compressing}
X.~Yu, T.~Liu, X.~Wang, and D.~Tao.
\newblock On compressing deep models by low rank and sparse decomposition.
\newblock In \emph{IEEE Conf. Comput. Vis. Pattern Recog.}, pages 7370--7379,
  2017.

\bibitem[Zha et~al.(2019)Zha, Yuan, Wen, Zhou, Zhang, and Zhu]{zha2019rank}
Z.~Zha, X.~Yuan, B.~Wen, J.~Zhou, J.~Zhang, and C.~Zhu.
\newblock From rank estimation to rank approximation: Rank residual constraint
  for image restoration.
\newblock \emph{IEEE Trans. Image Process.}, 29:\penalty0 3254--3269, 2019.

\bibitem[Zhan et~al.(2016)Zhan, Cao, Qian, Chang, and Wei]{zhan2016low}
M.~Zhan, S.~Cao, B.~Qian, S.~Chang, and J.~Wei.
\newblock Low-rank sparse feature selection for patient similarity learning.
\newblock In \emph{IEEE Int. Conf. on Data Min.}, pages 1335--1340. IEEE, 2016.

\bibitem[Zhang et~al.(2013{\natexlab{a}})Zhang, Ghanem, Liu, Xu, and
  Ahuja]{zhang2013low}
T.~Zhang, B.~Ghanem, S.~Liu, C.~Xu, and N.~Ahuja.
\newblock Low-rank sparse coding for image classification.
\newblock In \emph{Int. Conf. Comput. Vis.}, pages 281--288,
  2013{\natexlab{a}}.

\bibitem[Zhang et~al.(2013{\natexlab{b}})Zhang, Jiang, and
  Davis]{zhang2013learning}
Y.~Zhang, Z.~Jiang, and L.~S. Davis.
\newblock Learning structured low-rank representations for image
  classification.
\newblock In \emph{IEEE Conf. Comput. Vis. Pattern Recog.}, pages 676--683,
  2013{\natexlab{b}}.

\bibitem[Zou et~al.(2006)Zou, Hastie, and Tibshirani]{zou2006sparse}
H.~Zou, T.~Hastie, and R.~Tibshirani.
\newblock Sparse principal component analysis.
\newblock \emph{J. Comput. Graph. Stat.}, 15\penalty0 (2):\penalty0 265--286,
  2006.

\end{thebibliography}
}

\TOCstart
\renewcommand\thefigure{A\arabic{figure}}
\renewcommand\thetable{A\arabic{table}}  
\renewcommand\theequation{A\arabic{equation}}
\renewcommand\thealgorithm{A\arabic{algorithm}}

\appendix

\section*{Appendix}

\tableofcontents

\section{Proofs}
\subsection{Proof to \cref{corollary:rank theorem}}
This Lemma is the direct result of the \textit{rank theorem of manifolds}. 
\begin{theorem}[Rank Theorem~\cite{hirsch2012differential}]\label{th:rank_textbook}
Suppose $\f:\mathcal{M}\rightarrow\mathcal{N}$ is a smooth function from $m$-dimensional manifold $\mathcal{M}$ to $n$-dimensional manifold $\mathcal{N}$, and $\rank_{\mathcal{M},\mathcal{N}}(\bJ_{\f})=r$. Then for each $\x\in\mathcal{M}$, there exists a smooth chart $(\mathcal{U},\bm{m})$ around $\x$ and a smooth chart $(\mathcal{V},\bm{n})$ around $\f(\x)$, such that
\begin{equation}
    \bm{n}\circ\f\circ\bm{m}^{-1}:\bm{m}(\mathcal{U})\subset\mathbb{R}^m\rightarrow\bm{n}(\mathcal{V})\subset\mathbb{R}^n
\end{equation}
is given by $\bm{n}\circ\f\circ\bm{m}^{-1}(\x_1,\x_2,\cdots,\x_m)=(\x_1,\x_2,\cdots,\x_m,0,\cdots,0)$.
\end{theorem}
The rank of a function $\rank_{\mathcal{M},\mathcal{N}}$ defined on the manifold is the rank under local chart systems of the input manifold $\mathcal{M}$ and output manifold $\mathcal{N}$. Let $\bm{\phi}$ be the chart for point $\x\in\mathcal{O}\subset\mathcal{M}$, and the identity map be the chart for point $\f(\x)\in\mathbb{R}^d$. Then it is easy to find that
\begin{equation}
    \rank(\bm{I}^{-1}\circ\f\circ\bm{\phi})=\rank(\f\circ\bm{\phi})=\rank_{\mathcal{M},\mathcal{N}}(\f).
\end{equation}
Then by \cref{th:rank_textbook}, we know that
\begin{equation}
    \dim(\f(\mathcal{X}))=\rank_{\mathcal{M},\mathcal{N}}(\f)=\rank(\f\circ\bm{\phi})\leq\rank(\f)=r.
\end{equation}
The last equal sign comes from the rank inequality of matrix multiplication $\rank(\bm{A}\bm{B})\leq\min\{\rank(\bm{A}),\rank(\bm{B})\}$, which we will discuss later.

\subsection{Proof to \cref{th:nr_st}}
Proof to this Theorem needs  Weyl's inequalities~\cite{wely} for singular values of sum of matrices.
\begin{theorem}[Weyl's inequalities]
Let $\bA,\bB$ be $p\times n$ complex matrices, $\sigma_i(\cdot)$ be the $i$-th largest singular value of the matrix. Then
\begin{equation}
    \vert\sigma_i(\bA+\bB)-\sigma_i(\bB)\vert\leq\sigma_1(\bB),1\leq i\leq p,n.
\end{equation}
\end{theorem}
 Let $p$ be the number of singular values of $\bW$ and $\bD$. 
 By this theorem, we have
\begin{equation}
    \sigma_i(\bW)-\delta\sigma_1(\bD)\leq\sigma_i(\bW+\delta\bD)\leq \sigma_i(\bW)+\delta\sigma_1(\bD),i=1,\cdots, p.
\end{equation}
To measure the numerical rank, we need to estimate the relative quantities of singular values, which are,
\begin{equation}
    \frac{\sigma_i(\bW)-\delta\sigma_1(\bD)}{\sigma_1(\bW)+\delta \sigma_1(\bD)}\leq\frac{\sigma_i(\bW+\delta\bD)}{\sigma_1(\bW+\delta\bD)}\leq \frac{\sigma_i(\bW)+\delta\sigma_1(\bD)}{\sigma_1(\bW)-\delta \sigma_1(\bD)},i=2,\cdots,p.
\end{equation}
Now assume that $\epsilon$ does not belong to the following set (which is a zero measure set in $\mathbb{R}_+$)
\begin{equation}
    \Sigma_{\bW}=\left\{\frac{\sigma_i(\bW)}{\sigma_1(\bW)}:i=2,\cdots,p \right\},
\end{equation}
and $\rank_{\epsilon}(\bW)=r$. We know that
\begin{equation}
    \frac{\sigma_i(\bW)}{\sigma_1(\bW)}>\epsilon,i=2,\cdots,r;\frac{\sigma_i(\bW)}{\sigma_1(\bW)}<\epsilon,i=r+1,\cdots,p.
\end{equation}

Thus, we have that $\forall\delta<\delta_{\mathrm{max}}$,
\begin{gather}
    \frac{\sigma_i(\bW+\delta\bD)}{\sigma_1(\bW+\delta\bD)}\geq\frac{\sigma_i(\bW)-\delta\sigma_1(\bD)}{\sigma_1(\bW)+\delta \sigma_1(\bD)}>\epsilon,i=2,\cdots,r,\\
    \frac{\sigma_i(\bW)}{\sigma_1(\bW)}\leq\frac{\sigma_i(\bW)+\delta\sigma_1(\bD)}{\sigma_1(\bW)-\delta \sigma_1(\bD)}<\epsilon,i=r+1,\cdots,p,
\end{gather}
provided that
\begin{equation}\label{eq:sigma_max}
\begin{aligned}
        \delta_{\mathrm{max}}=\min\{\frac{1}{\sigma_1(D)}\left(\frac{\sigma_r(\bW)+\sigma_1(\bW)}{\epsilon+1}-\sigma_1(\bW)\right),\frac{\sigma_r(\bW)}{2\sigma_1(\bD)},\\
        \frac{1}{\sigma_1(D)}\left(\sigma_1(\bW)-\frac{\sigma_{r+1}(\bW)+\sigma_1(\bW)}{\epsilon+1}\right),\frac{\sigma_1(\bW)}{2\sigma_1(\bD)}\}.
\end{aligned}
\end{equation}
Thus we can conclude
\begin{equation}
    \rank_{\epsilon}(\bW+\delta\bD)=\rank_{\epsilon}(\bW),\forall\delta\in[0,\delta_{\mathrm{max}}).
\end{equation}
When $\rank(\bW)=r<p$, it is then easy to see if $\epsilon<\frac{\sigma_r(\bW)}{\sigma_1(\bW)}$, we have
\begin{gather}
    \frac{\sigma_i(\bW)}{\sigma_1(\bW)}>\epsilon,i=2,\cdots,r;\frac{\sigma_i(\bW)}{\sigma_1(\bW)}=0<\epsilon,i=r+1,\cdots,p.
\end{gather}
Thus setting $\epsilon_{\mathrm{max}}=\frac{\sigma_r(\bW)}{\sigma_1(\bW)}$, we can always have $\delta_{\mathrm{max}}$ acquired by \cref{eq:sigma_max}, such that
\begin{equation}
    \rank_{\epsilon}(\bW)=\rank(\bW)=\rank_{\epsilon}(\bW+\delta\bD),\forall\delta\in[0,\delta_{\mathrm{max}}).
\end{equation}

\subsection{Proof to \cref{th:partial rank}}
Let $\bA_i$ be the $i$-th column of matrix $\bA$. Given two matrices $\bA\in\mathbb{R}^{m\times n},\bB\in\mathbb{R}^{n\times d}$, we have
\begin{equation}
    \bA\bB=(\bA\bB_1,\cdots\bA\bB_d).
\end{equation}
Thus for any $1\leq i_1<\cdots<i_K\leq d$,
\begin{equation}
    (\bA\bB)_{i_1,\cdots,i_K}=(\bA\bB_{i_1},\cdots,\bA\bB_{i_K})=\bA(\bB)_{i_1,\cdots,i_K}.
\end{equation}
By the rank theorem~\cite{hoffman1971linear} of matrices, we have
\begin{equation}
    \rank((\bA\bB)_{i_1,\cdots,i_K})=\rank((\bB)_{i_1,\cdots,i_K})-\dim(\mathrm{Ker}(\bA)\cap\mathrm{Im}((\bB)_{i_1,\cdots,i_K})),
\end{equation}
and 
\begin{equation}
    \rank(\bA\bB)=\rank(\bB)-\dim(\mathrm{Ker}(\bA)\cap\mathrm{Im}(\bB).
\end{equation}
As $(\bB)_{i_1,\cdots,i_K})\subset \bB$, it is straightforward to get that
\begin{equation}
    \mathrm{Im}((\bB)_{i_1,\cdots,i_K}))\subset\mathrm{Im}(B).
\end{equation}
Thus
\begin{equation}
    \mathrm{Ker}(\bA)\cap\mathrm{Im}((\bB)_{i_1,\cdots,i_K})\subset\mathrm{Ker}(\bA)\cap\mathrm{Im}(\bB).
\end{equation}
Then we have
\begin{equation}
\begin{aligned}
           \rank(\bB)-\rank(\bA\bB)=\dim(\mathrm{Ker}(\bA)\cap\mathrm{Im}(\bB)\geq \dim(\mathrm{Ker}(\bA)\cap\mathrm{Im}((\bB)_{i_1,\cdots,i_K}))\\
           =\rank((\bB)_{i_1,\cdots,i_K}))-\rank((\bA\bB)_{i_1,\cdots,i_K})\geq 0. 
\end{aligned}
\end{equation}
Note that $\rank(\f_2\circ\f_1)=\rank(\bJ_{\f_2}\bJ_{\f_1})$. Then we complete the proof.
\subsection{Proof to \cref{th:principle}}
The key to this principle is the rank theorem of matrices~\cite{hoffman1971linear}, which is
\begin{equation}
    \rank(\bA\bB)=\rank(\bB)-\dim(\mathrm{Ker}(\bA)\cap\mathrm{Im}(\bB)).
\end{equation}
Note that $\rank(\bA\bB)=\rank(\bB^T\bA^T)$ and $\rank(\bA^T)=\rank(\bA)$. Then we have
\begin{equation}
\begin{aligned}
         \rank(\bA\bB)=\rank(\bB^T\bA^T)=\rank(\bA^T)-\dim(\mathrm{Ker}(\bB^T)\cap\mathrm{Im}(\bA^T))\\
         =\rank(\bA)-\dim(\mathrm{Ker}(\bB^T)\cap\mathrm{Im}(\bA^T)).      
\end{aligned}
\end{equation}
The dimension of a linear subspace will at least be zero, thus the above equations suggest
\begin{equation}
    \rank(\bA\bB)\leq\rank(\bB),\rank(\bA\bB)\leq\rank(\bA).
\end{equation}
Applying this argument to the chain rule of differentials then yields the conclusion. Further using \cref{corollary:rank theorem} gives the diminishing of intrinsic dimensions of feature manifolds.

\subsection{Proof to \cref{th:structural impetus}}
We first give the rigorous version of this theorem as follows.
\begin{theorem}
Let $\bm{e}_{\x}$ be the exponential map from a small neighborhood $\mathcal{U}_{\x}$ of point $\x$ on the input feature manifold to its tangent space at $\x$, and $\bm{v}_{\x}=\bm{e}_{\x}(\x)$. Let $\mathcal{X}$ be the input manifold, $s=\dim(\mathcal{X})$, $\f^i$ be the layer network, $r=\rank(\f^i)$, and $\f^i(\mathcal{X})$ be the output manifold. If for almost everywhere on the input feature manifold, there is a unit vector $\bm{v}\in\bm{e}_{\x}(\mathcal{U}_{\x})$, such that the layer network $\f^i$ satisfies
\begin{equation}\label{eq:exact_stru}
    \lim_{t\rightarrow0}\frac{\Vert\f^i\circ\bm{e}^{-1}_{\x}(\bm{v}_{\x})-\f^i\circ\bm{e}^{-1}_{\x}(\bm{v}_{\x}+t\bm{v})\Vert_2}{t}=0,
\end{equation}
then $\dim(\f^i(\mathcal{X}))<s$. If the number of such independent $\bm{v}$ in $\bm{e}_{\x}(\mathcal{U}_{\x})$ is $k$, then $\dim(\f^i(\mathcal{X}))\leq s-k$.
\end{theorem}
Now we prove this theorem. 

Note that \cref{eq:exact_stru} implies
\begin{equation}
    \bJ_{\bm{e}^{-1}_{\x}}\bm{v}\in \mathrm{Ker}(\bJ_{\f^i}).
\end{equation}
As it is also easy to see
\begin{equation}
    \bJ_{\bm{e}^{-1}_{\x}}\bm{v}\in\mathrm{Im}(\bJ_{\bm{e}^{-1}_{\x}}),
\end{equation}
we can conclude
\begin{equation}
    \bm{0}\neq \bJ_{\bm{e}^{-1}_{\x}}\bm{v}\in \mathrm{Ker}(\bJ_{\f^i})\cap\mathrm{Im}(\bJ_{\bm{e}^{-1}_{\x}}),
\end{equation}
where $0\neq\bJ_{\bm{e}^{-1}_{\x}}\bm{v} $ comes from the full rank property of exponential map and its inverse.
Thus we have
\begin{equation}
    \dim(\mathrm{Ker}(\bJ_{\f^i})\cap\mathrm{Im}(\bJ_{\bm{e}^{-1}_{\x}}))\geq 1.
\end{equation}

Specifically, if linearly independent $\bm{v}_1,\cdots,\bm{v}_k$ satisfy \cref{eq:exact_stru}, we can conclude
\begin{equation}
    \bm{0}\neq \bJ_{\bm{e}^{-1}_{\x}}\bm{v}_i\in \mathrm{Ker}(\bJ_{\f^i})\cap\mathrm{Im}(\bJ_{\bm{e}^{-1}_{\x}}),i=1,\cdots,k.
\end{equation}
As $\bJ_{\bm{e}^{-1}_{\x}}$ has full rank due to the property of exponential map, we know that $\bJ_{\bm{e}^{-1}_{\x}}\bm{v}_i,i=1,\cdots,k$ are linearly independent. Then
\begin{equation}
    \dim(\mathrm{Ker}(\bJ_{\f^i})\cap\mathrm{Im}(\bJ_{\bm{e}^{-1}_{\x}}))\geq k.
\end{equation}

Thus the rank theorem of matrices~\cite{hoffman1971linear} reads 
\begin{gather}
    \rank(\bJ_{\f^i\circ\bm{e}^{-1}_{\x}})=\rank(\bJ_{\f^i}\bJ_{\bm{e}^{-1}_{\x}})=\rank(\bJ_{\bm{e}^{-1}_{\x}})-\dim(\mathrm{Ker}(\bJ_{\f^i})\cap\mathrm{Im}(\bJ_{\bm{e}^{-1}_{\x}}))\\
    =s-\dim(\mathrm{Ker}(\bJ_{\f^i})\cap\mathrm{Im}(\bJ_{\bm{e}^{-1}_{\x}}))\leq s-k.
\end{gather}
Combining this result with \cref{th:rank_textbook} proves our result.

\subsection{Proof to \cref{th:general_deficiency}}
The proof to this theorem relies on the existence of Lyapunov exponents of dynamic systems. Given a linearized dynamic system
\begin{equation}
    \dot{\bm{v}}(t)=\bm{X}_t\bm{v},~\bm{v}(0)=\bm{v}_0\in\mathbb{R}^n,
\end{equation}
its (largest) Lyapunov exponent is defined as
\begin{equation}
    \lambda=\limsup_{t\rightarrow\infty}\frac{1}{t}\Vert \bm{v}\Vert_2.
\end{equation}
Further, for a sequence of subspace $\sL_h\subset \sL_{r-1}\subset\cdots\subset \sL_1\subset \sL_0=\mathbb{R}^n$, we can define the corresponding Lyapunov exponents of all those subspaces as
\begin{equation}
    \lambda_i=\lim_{t\rightarrow\infty}\frac{1}{t}\log\Vert \bm{v}\Vert_2,~ i=1,\cdots,h+1,~\bm{v}_0\in \sL_{i-1}\backslash \sL_{i},
\end{equation}
and we have
\begin{equation}
    \lambda=\lambda_1>\lambda_2>\cdots>\lambda_h.
\end{equation}
It may be surprising to find that such Lyapunov exponents exist, as $\bm{v}_0$ can traverse the entire subspace $\sL_{i-1}\backslash\sL_i$. We will demonstrate the existence of the Lyapunov exponents for our case later in \cref{sec:lyapunov_existence}, which is the classical results from the Furstenberg-Kesten theorem~\cite{furstenberg1960products} and multiplicative ergodic theorem~\cite{pesin1977characteristic}. Before that, we will first assume the existence of those Lyapunov exponents for simplicity of analysis.

Now consider the case of function couplings
\begin{equation}
    \F=\f^L\circ\cdots\circ\f^2\circ\f^1,
\end{equation}
which has the Jacobian matrix
\begin{equation}
    \bJ_{\F}=\bJ_{\f^L}\bJ_{\f^{L-1}}\cdots\bJ_{\f^2}\bJ_{\f^1}.
\end{equation}
Apparently, the following dynamic system induces the Jacobi matrix of $\F$,
\begin{equation}
    \dot{\bm{v}}(t)=\bJ_{\f^t}\bm{v},~\bm{v}(0)=\bm{v}_0\in\mathbb{R}^n,~t=1,\cdots,L.
\end{equation}
Thus its Lyapunov exponents are given by

\begin{equation}\label{eq:lyapunov_function_coupling}
    \lambda_i=\lim_{L\rightarrow\infty}\frac{1}{L}\log\Vert \bJ_{\F}\bm{v}_0\Vert_2, ~i=1,\cdots,h+1,~\bm{v}_0\in \sL_{i-1}\backslash \sL_{i},
\end{equation}
for a chain of subspaces $\{0\}=\sL_{h+1}\subset\sL_h\subset \sL_{h-1}\subset\cdots\subset \sL_1\subset \sL_0=\mathbb{R}^n$.

\subsubsection{Lyapunov exponents are limits of logarithms of subspace spectral norm divided by layer depth $L$}\label{sec:lyapunov_norm}
We first demonstrate that the Lyapunov exponents are limits of logarithm of the spectral norm of $\F$ on $\sL_{i-1}\backslash \sL_{i}$ divided by layer depth $L$ when $L\rightarrow\infty$, for $i=1,...,r$. 

It is easy to see
\begin{gather}
    \frac{1}{L}\log\Vert\bm{v}_0\Vert_2\Vert\bJ_{\F}\frac{\bm{v}_0}{\Vert\bm{v}_0\Vert_2}\Vert_2=\frac{1}{L}(\log\Vert\bm{v}_0\Vert_2+\log\Vert\bJ_{\F}\frac{\bm{v}_0}{\Vert\bm{v}_0\Vert_2}\Vert_2).
\end{gather}
When $L\rightarrow\infty$, $\frac{1}{L}\log\Vert\bm{v}_0\Vert_2\rightarrow0$ for any $\bm{v}_0$, we have
\begin{equation}\label{eq:lyapunov_singular}
    \lambda_i=\lim_{L\rightarrow\infty}\frac{1}{L}\log\Vert\bJ_{\F}\bm{v}_0\Vert_2,~ \Vert\bm{v}_0\Vert_2=1,~\bm{v}_0\in\sL_{i-1}\backslash\sL_i.
\end{equation}
Let 
\begin{equation}
    \lambda_i^L=\sup_{\Vert\bm{v}_0\Vert_2=1,\bm{v}_0\in\sL_{i-1}\backslash\sL_i}\frac{1}{L}\log\Vert\bJ_{\F}\bm{v}_0\Vert_2.
\end{equation}
Note that
\begin{equation}
    \sup_{\Vert\bm{v}_0\Vert_2=1,\bm{v}_0\in\sL_{i-1}\backslash\sL_i}\frac{1}{L}\log\Vert\bJ_{\F}\bm{v}_0\Vert_2=\frac{1}{L}\log\sup_{\Vert\bm{v}_0\Vert_2=1,\bm{v}_0\in\sL_{i-1}\backslash\sL_i}\Vert\bJ_{\F}\bm{v}_0\Vert_2,
\end{equation}
and 
\begin{equation}
    \sup_{\Vert\bm{v}_0\Vert_2=1,\bm{v}_0\in\sL_{i-1}\backslash\sL_i}\Vert\bJ_{\F}\bm{v}_0\Vert_2=\Vert\bJ_{\F}\Vert_{2,i},
\end{equation}
where $\Vert\cdot\Vert_{2,i}$ denote the spectral norm of a linear operator constrained on $\sL_{i-1}\backslash\sL_i$.Then we have
\begin{equation}
    \lambda_i^L=\frac{1}{L}\log\Vert\bJ_{\F}\Vert_{2,i}.
\end{equation}

Let $\bm{e}_{i_1},\cdots,\bm{e}_{i_k}$ be a set of standard orthogonal basis of $i_k$ dimensional subspace $\sL_{i-1}\backslash\sL_i$. If the Lyapunov exponents exist, by \cref{eq:lyapunov_singular} we have for any $\epsilon>0$, there is $N\in\mathbb{N}$ such that for all $L>N$,
\begin{equation}\label{eq:sup_lyapunov_singular}
    \lambda_i-\frac{\epsilon}{2}\leq\frac{1}{L}\log\Vert \bJ_{\F}\bm{e}_{j}\Vert_2\leq \lambda_i+\frac{\epsilon}{2},j=1,\cdots,i_k.
\end{equation}
Let $\bm{v}=\sum_{j=1}^{i_k}\alpha_j\bm{e}_{j}\in\sL_{i-1}\backslash\sL_i$, where $\alpha_j,i=1,\cdots,i_k,\sum_{j=1}^{i_k}\alpha_j^2=1$ is the coordinate of unit vector $\bm{v}$ under the basis $\bm{e}_{j},j=1,\cdots,i_k$. Assume that $\Vert\bJ_{\F}\bm{e}_1\Vert_2\geq\Vert\bJ_{\F}\bm{e}_j\Vert_2,j=2,\cdots,i_k$. We then have $\frac{\vert\alpha_j\vert\Vert\bJ_{\F}\bm{e}_j\Vert_2}{\sum_{j=1}^{i_k}\vert\alpha_j\vert\Vert\bJ_{\F}\bm{e}_1\Vert_2}\leq1,j=1,\cdots,i_k$, and
\begin{gather}
    \frac{1}{L}\log\Vert\bJ_{\F}\bm{v}\Vert_2
    \leq\frac{1}{L}\log\sum_{j=1}^{i_k}\vert\alpha_j\vert \Vert\bJ_{\F}\bm{e}_{j}\Vert_2\\
    =\frac{1}{L}\log\left(\frac{\vert\alpha_1\vert\Vert\bJ_{\F}\bm{e}_1\Vert_2}{\sum_{j=1}^{i_k}\vert\alpha_j\vert\Vert\bJ_{\F}\bm{e}_1\Vert_2}+\sum_{j=2}^{i_k}\frac{\vert\alpha_j\vert \Vert\bJ_{\F}\bm{e}_{j}\Vert_2}{\sum_{j=1}^{i_k}\vert\alpha_j\vert\Vert\bJ_{\F}\bm{e}_1\Vert_2}\right)(\sum_{j=1}^{i_k}\vert\alpha_j\vert\Vert\bJ_{\F}\bm{e}_1\Vert_2)\\
    =\frac{1}{L}\log\left(\frac{\vert\alpha_1\vert\Vert\bJ_{\F}\bm{e}_1\Vert_2}{\sum_{j=1}^{i_k}\vert\alpha_j\vert\Vert\bJ_{\F}\bm{e}_1\Vert_2}+\sum_{j=2}^{i_k}\frac{\vert\alpha_j\vert \Vert\bJ_{\F}\bm{e}_{j}\Vert_2}{\sum_{j=1}^{i_k}\vert\alpha_j\vert\Vert\bJ_{\F}\bm{e}_1\Vert_2}\right) +\frac{1}{L}\log\sum_{j=1}^{i_k}\vert\alpha_j\vert\Vert\bJ_{\F}\bm{e}_1\Vert_2 \\
    \leq\frac{1}{L}\log\left(1+\sum_{j=2}^{i_k}\frac{\vert\alpha_j\vert \Vert\bJ_{\F}\bm{e}_{j}\Vert_2}{\sum_{j=1}^{i_k}\vert\alpha_j\vert\Vert\bJ_{\F}\bm{e}_1\Vert_2}\right) +\frac{1}{L}\log\sum_{j=1}^{i_k}\vert\alpha_j\vert\Vert\bJ_{\F}\bm{e}_1\Vert_2 \\
    \leq\frac{1}{L}\sum_{j=2}^{i_k}\frac{\vert\alpha_j\vert \Vert\bJ_{\F}\bm{e}_{j}\Vert_2}{\sum_{j=1}^{i_k}\vert\alpha_j\vert\Vert\bJ_{\F}\bm{e}_1\Vert_2} +\frac{1}{L}\log\sum_{j=1}^{i_k}\vert\alpha_j\vert\Vert\bJ_{\F}\bm{e}_1\Vert_2 \\
    \leq \frac{1}{L}(i_k-1)+\frac{1}{L}\log\sum_{j=1}^{i_k}\vert\alpha_j\vert +\frac{1}{L}\log\Vert\bJ_{\F}\bm{e}_1\Vert_2\\
    \leq \frac{1}{L}(i_k-1)+\frac{1}{2L}\log(1^2+\cdots+1^2)(\alpha_1^2+\cdots+\alpha_{i_k}^2)+\frac{1}{L}\log\Vert\bJ_{\F}\bm{e}_1\Vert_2\\
    =\frac{1}{L}(i_k-1)+\frac{1}{2L}\log i_k+\frac{1}{L}\log\Vert\bJ_{\F}\bm{e}_1\Vert_2\\
    \leq \frac{1}{L}(i_k-1)+\frac{1}{2L}\log i_k+\lambda_i+\frac{\epsilon}{2}.
\end{gather}
 Thus, if we set $N_0=\max\{N, \frac{2i_k-2+\log i_k}{\epsilon}\}$, then when $L>N_0$ we have $\frac{1}{L}(i_k-1)+\frac{1}{2L}\log i_k<\frac{\epsilon}{2}$ and 
\begin{equation}\label{eq:all_unit_lyapunov}
    \frac{1}{L}\log\Vert\bJ_{\F}\bm{v}\Vert_2\leq \lambda_i+\epsilon,~\forall\bm{v}\in\sL_{i-1}\backslash\sL_i,~\Vert\bm{v}\Vert_2=1.
\end{equation}
Combining \cref{eq:sup_lyapunov_singular,eq:all_unit_lyapunov}, we have for any $\epsilon>0$, there is $N_0\in\mathbb{N}$, such that when $L>N_0$, we always have
\begin{equation}
    \lambda_i-\frac{\epsilon}{2}\leq\lambda_i^L=\sup_{\Vert\bm{v}_0\Vert_2=1,\bm{v}_0\in\sL_{i-1}\backslash\sL_i}\frac{1}{L}\log\Vert\bJ_{\F}\bm{v}_0\Vert_2=\frac{1}{L}\log\Vert\bJ_{\F}\Vert_{2,i}\leq \lambda_i+\epsilon.
\end{equation}

Thus, if the Lyapunov exponents exist, \ieno, the existence of limits of \cref{eq:lyapunov_function_coupling} , we have
\begin{equation}
    \lambda_i=\lim_{L\rightarrow\infty}\frac{1}{L}\log\Vert\bJ_{\F}\Vert_{2,i}=\lim_{L\rightarrow\infty}\lambda_i^L.
\end{equation}


\subsubsection{Singular value distributions of Jacobian matrices of deep function coupling}
In \cref{sec:lyapunov_norm} we have proved that the Lyapunov exponents (if they exist) are limits of logarithms of subspace spectral norms divided by $L$. Here we use this property to prove the deficiency of numerical ranks, \ieno, \cref{eq:implicit_impetus}.

We first introduce the Courant-Fischer min-max theorem~\cite{hoffman1971linear} of sigular values.
\begin{theorem}[Courant-Fischer Min-max Theorem]\label{th:minmax}
Let $\bA$ be a $d\times n$ complex matrix and $\sigma_i(\bA)$ denote its $i$-th largest singular value, $i=1,\cdots,\min\{d,n\}$. Then we have
\begin{gather}
\sigma_i(\bA)=\sup_{\dim(\mathcal{V})=i}\inf_{\bm{v}\in\mathcal{V},\Vert\bm{v}\Vert_2=1}\Vert \bA\bm{v}\Vert_2,\\
\sigma_i(\bA)=\inf_{\dim(\mathcal{V})=n-i+1}\sup_{\bm{v}\in\mathcal{V},\Vert\bm{v}\Vert_2=1}\Vert \bA\bm{v}\Vert_2,\\
\end{gather}
where $\mathcal{V}$ traverses subspaces of $\mathbb{R}^n$.
\end{theorem}
This theorem also serves as one of the definitions to singular values.

Now assume that $\dim(\sL_0\backslash\sL_1)=r$, and we consider only the case of $d=n$ for simplicity. For any $\epsilon>0$, we have $N\in\mathbb{N}$ such that when $L>N$, 
\begin{gather}
    \inf_{\bm{v}\in\sL_0\backslash\sL_1,\Vert\bm{v}\Vert_2=1}\Vert\bJ_{\F}\bm{v}\Vert_2\geq \exp L(\lambda_1-\epsilon)\\
\end{gather}
due to \cref{eq:lyapunov_singular}. As $\dim(\sL_0\backslash\sL_1)=r$, by \cref{th:minmax}, we have
\begin{equation}\label{eq:lambda1}
    \sigma_1(\bJ_{\F})\geq\cdots \geq\sigma_r(\bJ_{\F})\geq \exp L(\lambda_1-\epsilon).
\end{equation}

For $\bm{v}\in\sL_0=\mathbb{R}^n$ and $\Vert\bm{v}\Vert_2=1$, as $\sL_0=\sL_0\backslash\sL_1\oplus\cdots\oplus\sL_{h}\backslash\sL_{h+1}$ ($\oplus$ denotes direct sum of linear quotient subspaces in the Banach space), there is $\bm{v}_i\in\sL_{i-1}\backslash\sL_{i},i=1,\cdots,h+1$, such that
\begin{equation}
    \Vert\bm{v}_1\Vert_2^2+\cdots +\Vert\bm{v}_{h+1}\Vert_2^2=1
\end{equation}
and
\begin{equation}
    \bm{v}=\bm{v}_1+\cdots +\bm{v}_{h+1}.
\end{equation}
Then by the conclusion of \cref{sec:lyapunov_norm}, we have
\begin{gather}
    \limsup_{L\rightarrow\infty}\Vert\bJ_{\F}\bm{v}\Vert_2\leq\limsup_{L\rightarrow\infty}\Vert\bJ_{\F}\bm{v}_1\Vert_2+\cdots+\limsup_{L\rightarrow\infty}\Vert\bJ_{\F}\bm{v}_{h+1}\Vert_2\\
    \leq \Vert\bm{v}_1\Vert_2\lim_{L\rightarrow\infty}\Vert\bJ_{\F}\Vert_{2,2}+\cdots+\Vert\bm{v}_{h+1}\Vert_2\lim_{L\rightarrow\infty}\Vert\bJ_{\F}\Vert_{2,h+1}
    \leq\sum_{i=1}^{h+1}\Vert\bm{v}_i\Vert_2\lambda_i\leq \lambda_1.
\end{gather}
Thus there is $N_1\in\mathbb{N}$, such that when $L>N_1$, we have
\begin{equation}
    \sup_{\bm{v}\in\sL_0,\Vert\bm{v}\Vert_2=1}\Vert\bJ_{\F}\bm{v}\Vert_2\leq\lambda_1+\epsilon.
\end{equation}

As $\dim(\sL_1)=n-1+1$, by \cref{th:minmax}, we have when $L>N_1$,
\begin{equation}
    \sigma_r(\bJ_{\F})\leq\cdots\leq\sigma_1(\bJ_{\F})\leq\exp L(\lambda_1+\epsilon).
\end{equation}

In conclusion, when $L>N_0=\max\{N,N_1\}$, we have
\begin{equation}
    \exp L(\lambda_1-\epsilon)\leq\sigma_1(\bJ_{\F})\leq\exp L(\lambda_1+\epsilon).
\end{equation}
Thus when $L\rightarrow\infty$, we have
\begin{equation}
    \sigma_1(\bJ_{\F})\sim \exp L\lambda_1.
\end{equation}

Using the same argument for $\sigma_2(\bJ_{\F}),\cdots,\sigma_n(\bJ_{\F})$, we can find that if let $\hat{\lambda}_1\geq\hat{\lambda}_2\cdots\geq\hat{\lambda}_n$ be the Lyapunov exponents counting repetitions, \ieno,
\begin{equation}
    \hat{\lambda}_{k}=\lambda_i,\text{ if } \sum_{j=1}^{i-1}\dim(\sL_{j-1}\backslash\sL_j)<k\leq\sum_{j=1}^i\dim(\sL_{j-1}\backslash\sL_j),i=1,\cdots,h+1,
\end{equation}
then
\begin{equation}
    \sigma_i(\bJ_{\F})\sim \exp L\hat{\lambda}_i.
\end{equation}

Note that 
\begin{equation}
    \hat{\lambda}_1=\cdots=\hat{\lambda}_r=\lambda_1,\hat{\lambda}_i\leq\lambda_2<\lambda_1,i=r+1,\cdots,n.
\end{equation}

Thus we have
\begin{equation}
    \frac{\sigma_i(\bJ_{\F})}{\sigma_1(\bJ_{\F})}\sim \exp L(\hat{\lambda}_i-\hat{\lambda}_1)\rightarrow0,i=r+1,\cdots,n.
\end{equation}

As a consequence, $\rank_{\epsilon}(\F)\leq r$ for any $\epsilon>0$ when $L\rightarrow\infty$.

\subsubsection{Existence of Lyapunov exponents for Jacobian matrices of deep function coupling}\label{sec:lyapunov_existence}
In above analysis, we have proven \cref{th:general_deficiency} under the existence of Lyapunov exponents. In this section, we introduce the classical result of multiplicatve ergodic theorem in the specific domain of random matrices, which is proposed by Furstenberg and Kesten~\cite{furstenberg1960products,FKT}.

\begin{theorem}[Multiplicatve Ergodic Theorem (Theorem 3.9 of~\cite{FKT})]\label{th:met}
Let $\mu$ be a probability measure on all convertible matrices of $\mathbb{R}^{n\times n}$ which satisfies 
\begin{equation}
    \mathbb{E}_{\mu}[\max\{\log\Vert\bJ_{\f^k}^{\pm1}\Vert_2,0\}]<\infty,~k=1,\cdots,L.
\end{equation}
If each $\bJ_{\f^k}$ independently follows $\mu$, then we have a chain of subspaces $\{0\}=\sL_{h+1}\subset\sL_h\subset\cdots\subset\sL_1\subset\sL_0=\mathbb{R}^n$ and corresponding postive real constants $\lambda_1>\lambda_2>\cdots>\lambda_{h+1}$ such that almost surely
\begin{equation}
    \lambda_i = \lim \frac{1}{t}\log\Vert\bJ_{\F}\bm{v}\Vert_2,~\forall\bm{v}\in\sL_{i-1}\backslash\sL_i,~i=1,\cdots,h+1,
\end{equation}
which means the existence of the Lyapunov exponents.
\end{theorem}
Combining this theorem and the arguments above, we can finally prove \cref{th:general_deficiency}.

\subsection{Proof to \cref{th:lyapunov_gaussian}}
This theorem can be deduced from the Lyapunov components of Ginibre matrices (polynomial ensemble of square matrices sampled \textit{i.i.d} from standard Gaussian).
\begin{theorem}[Exact Lyapunov Exponent Distribution for Ginibre Matrices~\cite{newman1986distribution}]\label{th:newman}
If $\mu$ in \cref{th:met} is standard Gaussian, then $h+1=n$, and
\begin{equation}
    \lambda_i=\log\left(2+\psi(\frac{n-i+1}{2})\right),i=1,\cdots,n.
\end{equation}
\end{theorem}
Combining this theorem with \cref{th:general_deficiency} can directly yield our result.

\subsection{Proof to \cref{corollary:ge}}
This theorem is the direct result of \cref{th:general_deficiency,th:newman}. Note that it is easy to get 
\begin{equation}
    \lambda_1=\lim_{L\rightarrow\infty}\frac{1}{L}\log\Vert\bJ_{\F}\Vert_2
\end{equation}
for standard Gaussian $\mu$.
\section{Possible Remission Approaches to Rank Diminishing}
\paragraph{Skip Connection}
Skip Connection is the most direct method to solve rank diminishing. 
In our formulation, the definition of a layer network requires it to accept inputs purely from its predecessor layer as 
\begin{equation}
    \x^{i}=\f^i(\x^{i-1}),~\x^{i-1}=\f^{i-1}(\x^{i-2}).
\end{equation}
However, when we add a skip connection from its ancestor layer $\f^s,s<i-1$, we have
\begin{equation}
    \x^i=\f^i(\x^{i-1},\x^s),\x^{i-1}=\f^{i-1}(\x^{i-2}),\x^{i-2}=\f^{i-2}\circ\cdots\circ\f^s(\x^{s-1}),\x^s=\f^s(\x^{s-1}).
\end{equation}
It actually makes the coupling of layers
\begin{equation}
    \hat{\f}^s=\left(\begin{array}{c}
    \f^{i-1}\circ\cdots \circ\f^s\\
    \f^s
    \end{array}\right),
\end{equation}
the true predecessor layer to $\f^i$, as 
\begin{equation}
    \x^{i}=\f^i(\hat{\x}),~\hat{\x}=\hat{f}^{s}(\x^{s-1}).
\end{equation}
Thus the true layer depth is cut down by $i-s$, remaining $L-(i-s)$ layers. Skip connection is usually used with the residual network. This structure can ease rank diminishing inside the layer $\hat{f}^s$, which we will discuss later. Overall, skip connection shortens the length of the chain of Jacobian matrices, thus restraining  rank diminishing.

\paragraph{BatchNorm}
Some previous works~\cite{BNpreventRC,understandingBN} discuss the role of BatchNorm in restraining rank diminishing. They show that BatchNorm may slow down the speed of rank diminishing in neural networks in some specific cases.

\paragraph{Residual Network}
Residual Network is another useful tool to restrain rank diminishing. The residual network $\bm{r}$ has the form
\begin{equation}
    \x^o=\bm{r}(\x^i)=\x^i+\mathrm{Res}(\x^i),
\end{equation}
where $\x^o$ and $\x^i$ are the output feature and input feature, respectively. Usually the residual term $\mathrm{Res}(\x^i)$ is small compared with the input $\x^i$. Assume that
\begin{equation}
    \Vert\bJ_{\mathrm{Res}}\Vert_2<\epsilon,
\end{equation}
where $\epsilon$ is very small. Then we have
\begin{equation}
    \bJ_{\bm{r}}=\bm{I}+\bJ_{\mathrm{Res}}
\end{equation}
is a diagonally dominant matrix, thus it has full rank. This means its kernel space $\mathrm{Ker}(\bJ_{\bm{r}})=\{0\}$ is a zero dimension space. Thus by the Rank Theorem, for any predecessor layer $\f$, $\bm{r}\circ\f$ will not lose rank as
\begin{equation}
\begin{aligned}
        \rank(\bm{r}\circ\f)=\rank(\bJ_{\bm{r}}\bJ_{\f})=\rank(\bJ_{\f})-\dim(\mathrm{Ker}(\bJ_{\bm{r}})\cap\mathrm{Im}(\bJ_{\f}))\\
        =\rank(\bJ_{\f})=\rank(\f).
\end{aligned}
\end{equation}
\section{Code}

\cref{alg:PartialRank} provides the pseudo-code of partial rank of the Jacobian. The implementation of the ~\cref{alg:PartialRank} can refer to the `rank\_jacobian.py' python file.

\cref{alg:NumericalRank} provides the pseudo-code of perturbed PCA dimension of feature spaces. The implementation of the \cref{alg:NumericalRank} can refer to the `rank\_perturb.py' python file.

\cref{alg:IntrinsicDimension} provides the pseudo-code of the classification dimension. The implementation of the~\cref{alg:IntrinsicDimension} can refer to the `run\_cls\_dim.py' python file.

\cref{alg:Hamster} provides the pseudo-code of independence deficit. The implementation of the~\cref{alg:Hamster} can refer to the `run\_deficit.py' python file.


\section{Partial Rank of Jacobians under Different Input Patches}
\begin{figure}[htbp]
    \centering
    \includegraphics[width=1.0\linewidth]{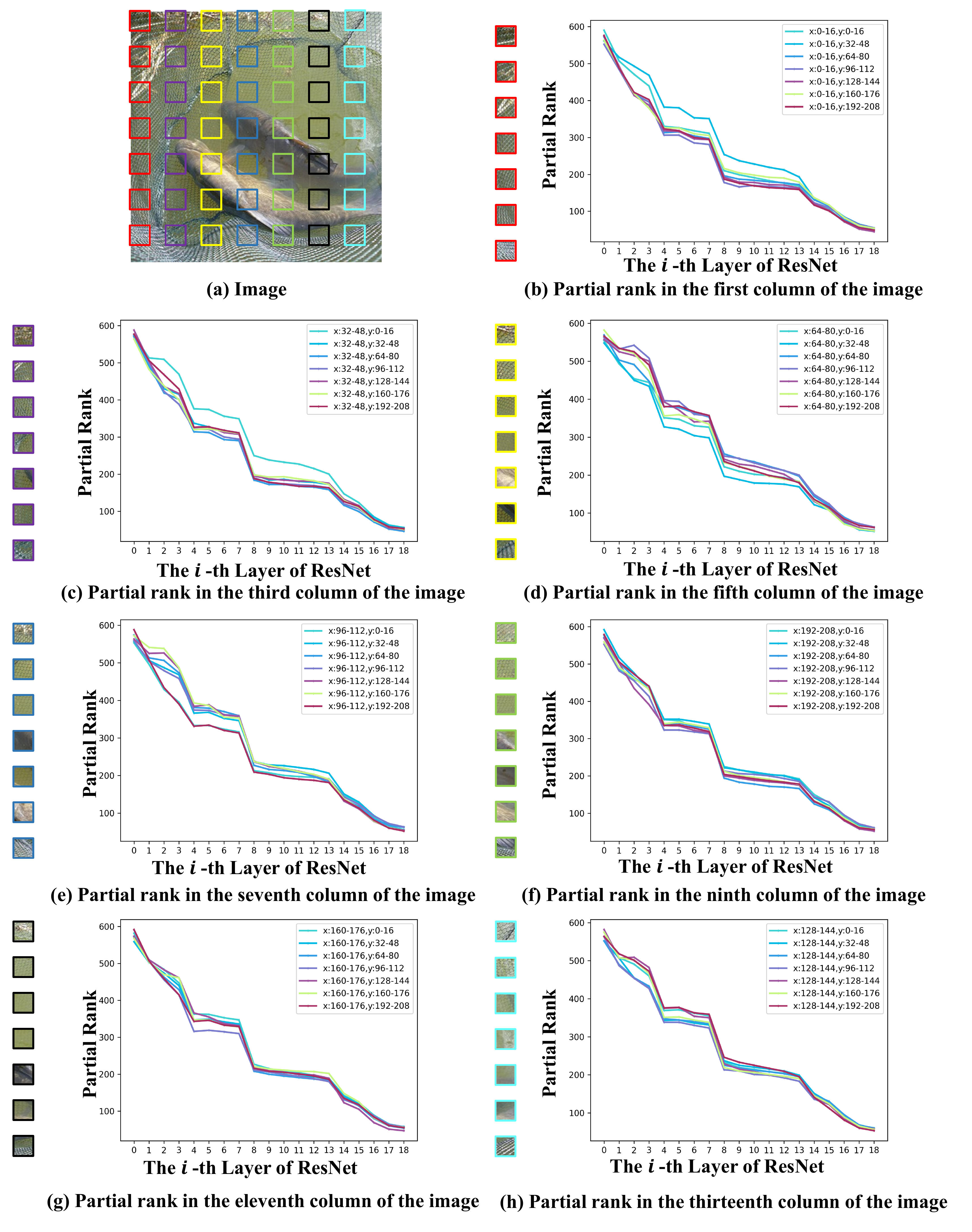}
    \caption{The partial ranks of different input patches at the $i$-th layer of ResNet-50 on ImageNet.}
    \label{fig:exp_supp_jac_rank_patch}
\end{figure}

In~\cref{fig:exp_supp_jac_rank_patch} we report partial ranks of different input image patches (marked with colored boxes in~\cref{fig:exp_supp_jac_rank_patch}(a)) for the layers of ResNet-50 on ImageNet. We can find that the curves of partial ranks share a similar and consistent trend among different input patches. Thus, picking one patch, for example, the central patch of $16\times16\times3$ pixels we use in \cref{Validations}, could be enough to demonstrate the overall behavior of network ranks. The consistent behavior of all those partial ranks also shows that partial rank is a good tool to investigate network ranks.




\section{Estimating Dimension Diminishing in Features}
Measuring the intrinsic dimension of feature manifolds is known to be hard. However, we manage to give a rough estimation to the dimension dropped by different layer networks. To do this, we use a new metric called the Perturbed PCA Dimension. It measures the expectation of PCA dimension of small local neighborhoods over the feature manifold.

Let $\F_k$ be the $k$-th sub-network of the whole network $\F$. We want to measure the Perturbed PCA Dimension of $\F_k(\mathcal{X})$, where $\mathcal{X}$ is the input data domain. To this end, we compute
\begin{equation}
    \mathrm{PertDim}=\mathbb{E}_{\x\sim\mathbb{P}_{\mathcal{X}}}[\mathrm{PCADim}(\{\F_k(\x+\bm{\epsilon}):\bm{\epsilon}\sim\mathcal{N}(\mathcal{O},\delta\mathcal{I})\})],
\end{equation}
where $\mathrm{PCADim}$ for a set is the number of PCA eigenvalues larger than a threshold $\xi$. For each point $\x$, we sample 50,000 different perturbation $\bm{\epsilon}$ to compute the PCA dimension of the neighborhood of $\F_k(\x)$. When computing the PCA dimension, we set $\delta=1e-3$ and $\xi=1.19e-7\times 50000\times \mathrm{eig}_{\mathrm{max}}$, where $\mathrm{eig}_{\mathrm{max}}$ is the largest PCA eigenvalue. We then compute the mean value of PCA dimensions over the neighborhood of 100 random samples in the validation set of ImageNet as the final result.

 We do not use PCA dimension of the feature manifolds directly as it is unable to cope with the highly non-linear structure of intermediate feature manifolds. However, the Perturbed PCA Dimension is able to estimate the dimensions of local neighborhoods of points in the feature manifolds. As local neighborhoods can be viewed as linear if the network is smooth, the Perturbed PCA Dimension could be more feasible than PCA dimension in our case. We provide the pseudo-code to compute the Perturbed PCA Dimension in \cref{alg:NumericalRank}.
 
\begin{figure}[htbp]
    \centering
    \includegraphics[width=1.0\linewidth]{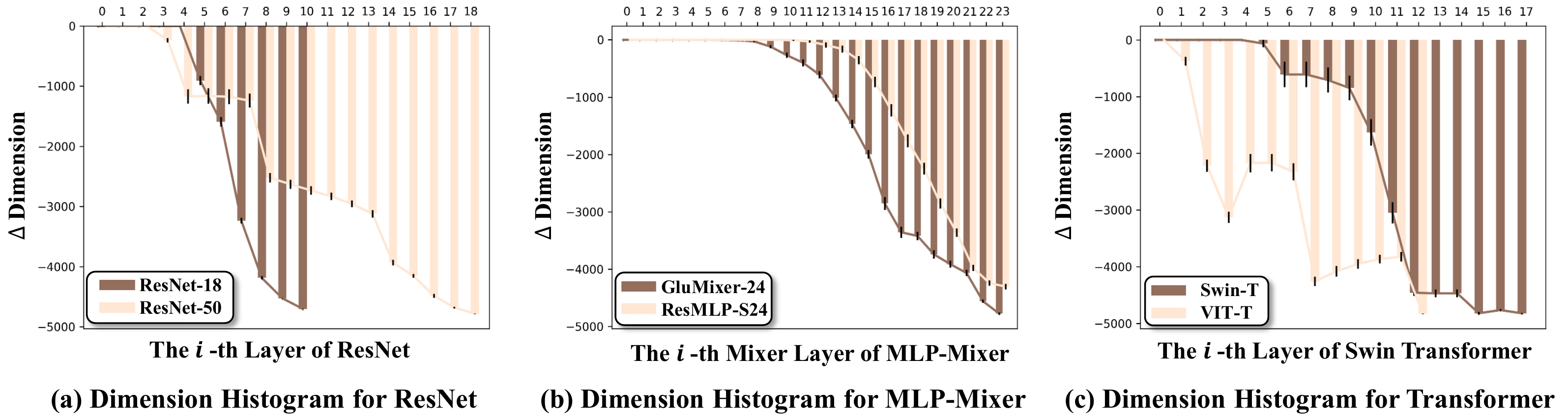}
    \caption{Dropped Perturbed PCA Dimension of different layers. $\Delta$ Dimension for the $i$-th layer is the difference between the Perturbed PCA Dimension of the $i$-th layer and that of the input layer of the CNN, MLP, and Transformer architectures on ImageNet.}
    \label{fig:exp_supp_fea_rank}
\end{figure}

 However, the perturbation is made in the ambient space of the input data manifold $\mathcal{X}$ rather than the data manifold itself. Thus this estimation may considerably overestimate the intrinsic dimensions of feature manifolds. So we merely care about how many Perturbed PCA Dimensions are lost by a sub-network instead of its own Perturbed PCA Dimension. We call this quantity $\Delta$ Dimension, which is the difference between the Perturbed PCA Dimension of the current layer and that of the input layer for the given deep network. As shown in \cref{fig:exp_supp_fea_rank}, we show the dropped dimensions of different feature layers of the CNN, MLP, and Transformer architectures on ImageNet. The results show that the Perturbed PCA Dimensions of feature manifolds of most networks decrease as the networks get deeper, thus confirming the rank diminishing principle we propose in \cref{th:principle}.

\section{More Examples of Independence Deficit}
\begin{figure}[t]
\centering
\includegraphics[width=0.98\linewidth]{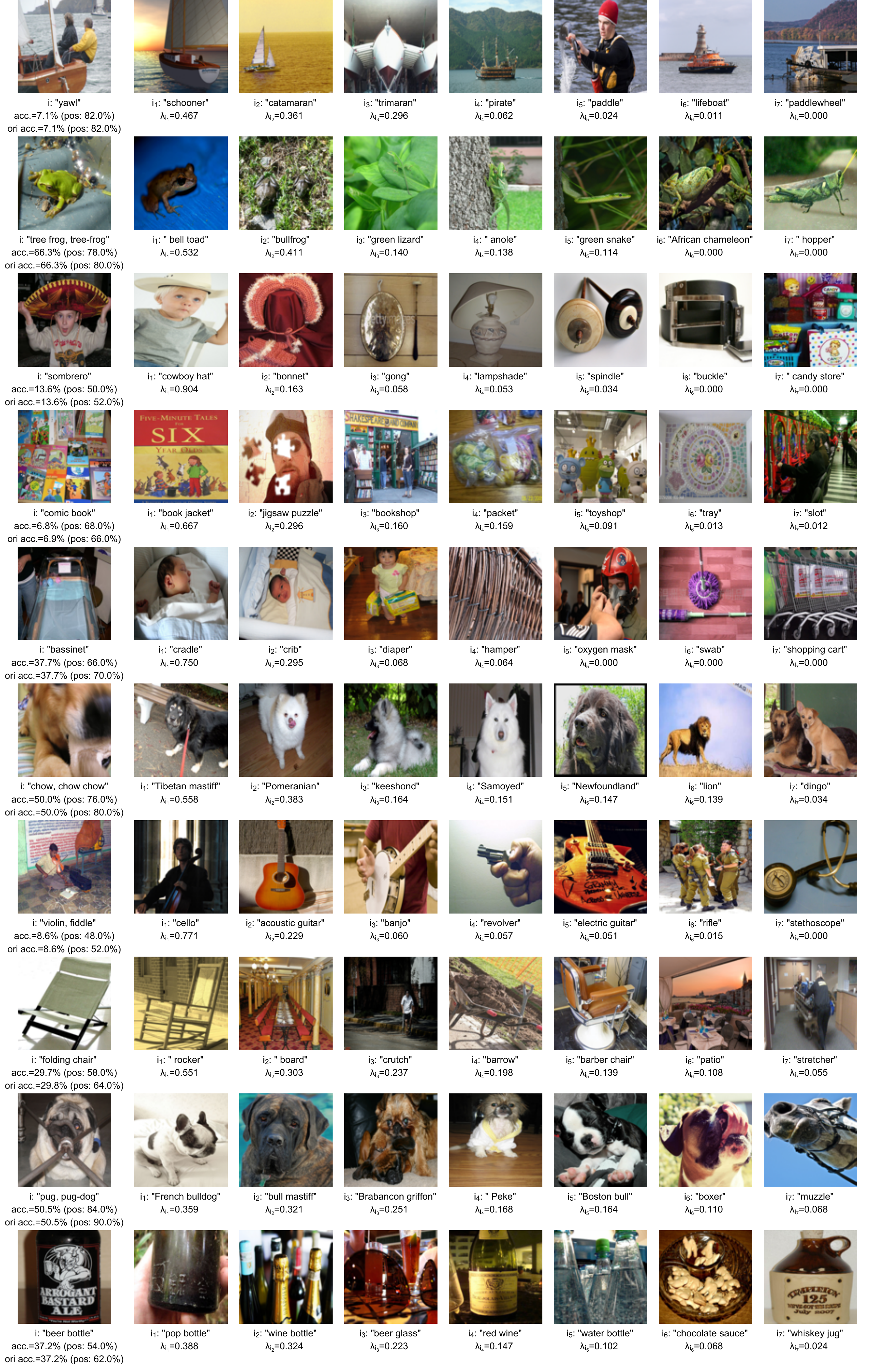}
\caption{Results from ResNet-18, where `acc.' and `ori acc.' denote the classification accuracies on the ImageNet validation set, while `pos: xx\%' is the accuracy on positive samples only.}
\label{fig:supp_deficit_res18}
\end{figure}

\begin{figure}[t]
\centering
\includegraphics[width=0.98\linewidth]{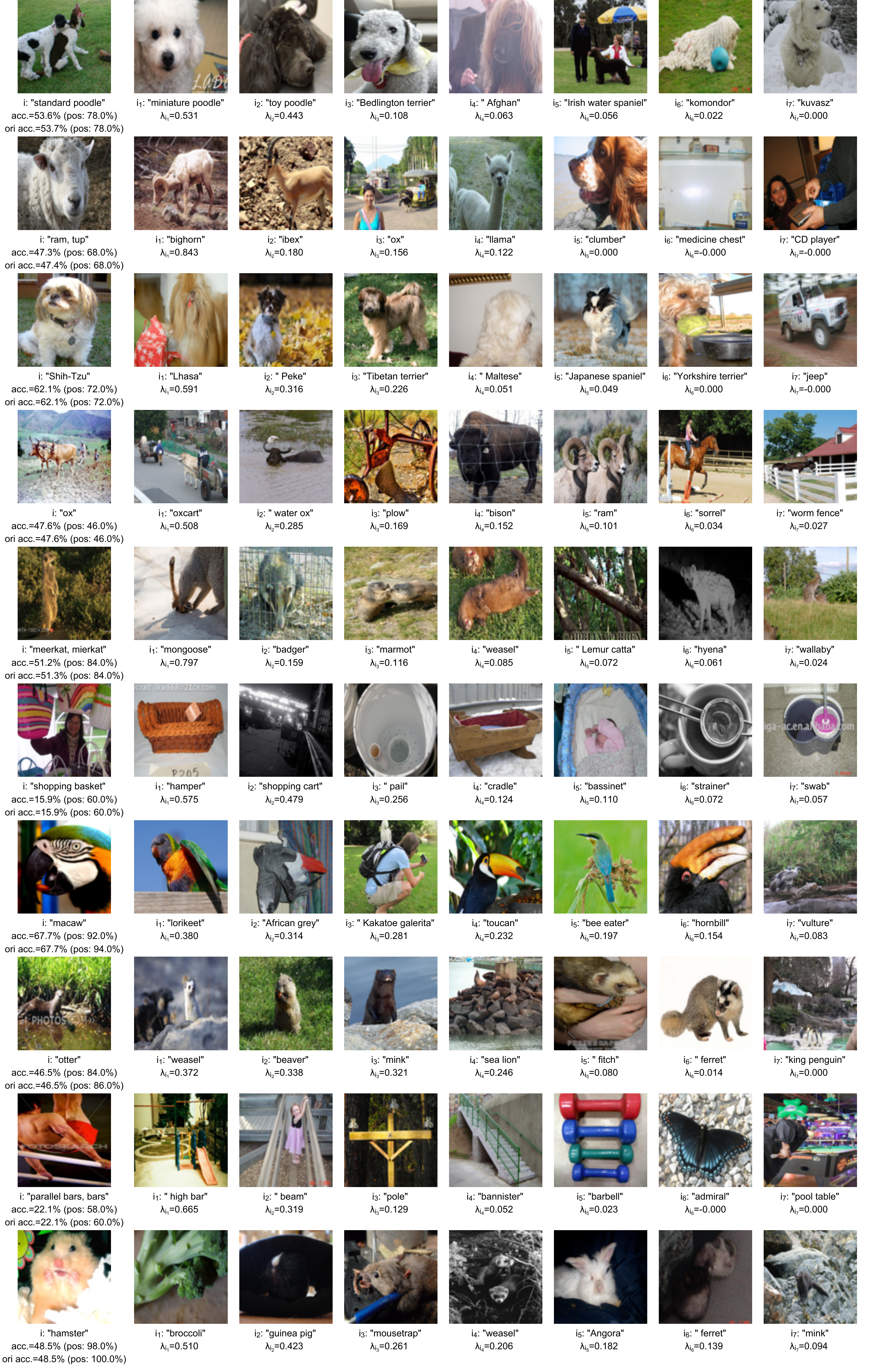}
\caption{Results from ResNet-50, where `acc.' and `ori acc.' denote the classification accuracies on the ImageNet validation set, while `pos: xx\%' is the accuracy on positive samples only.}\label{fig:supp_deficit_res50}
\end{figure}

\begin{figure}[t]
\centering
\includegraphics[width=0.98\linewidth]{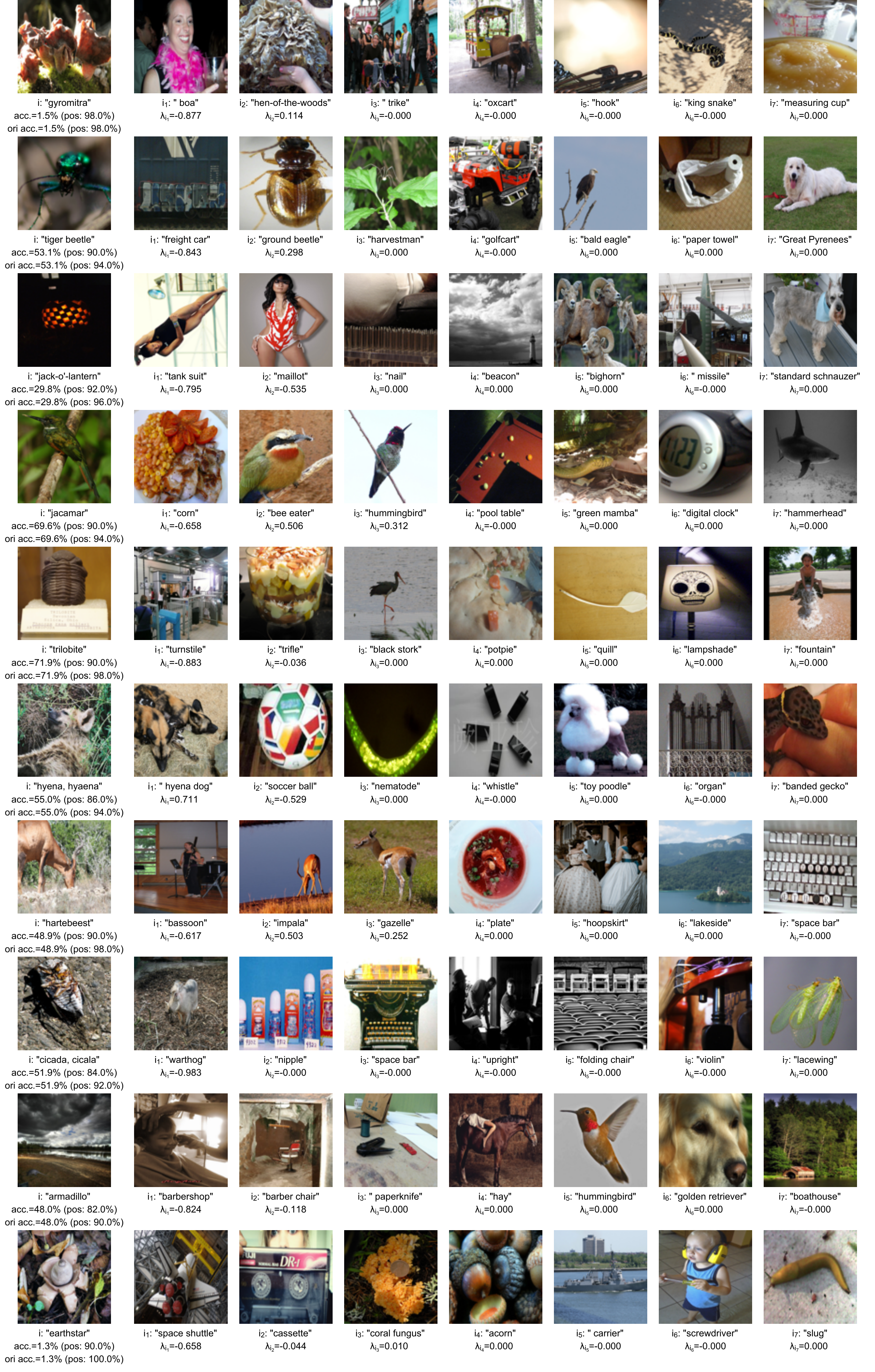}
\caption{Results from GluMixer-24, where `acc.' and `ori acc.' denote the classification accuracies on the ImageNet validation set, while `pos: xx\%' is the accuracy on positive samples only.}\label{fig:supp_deficit_gmlp}
\end{figure}

\begin{figure}[t]
\centering
\includegraphics[width=0.98\linewidth]{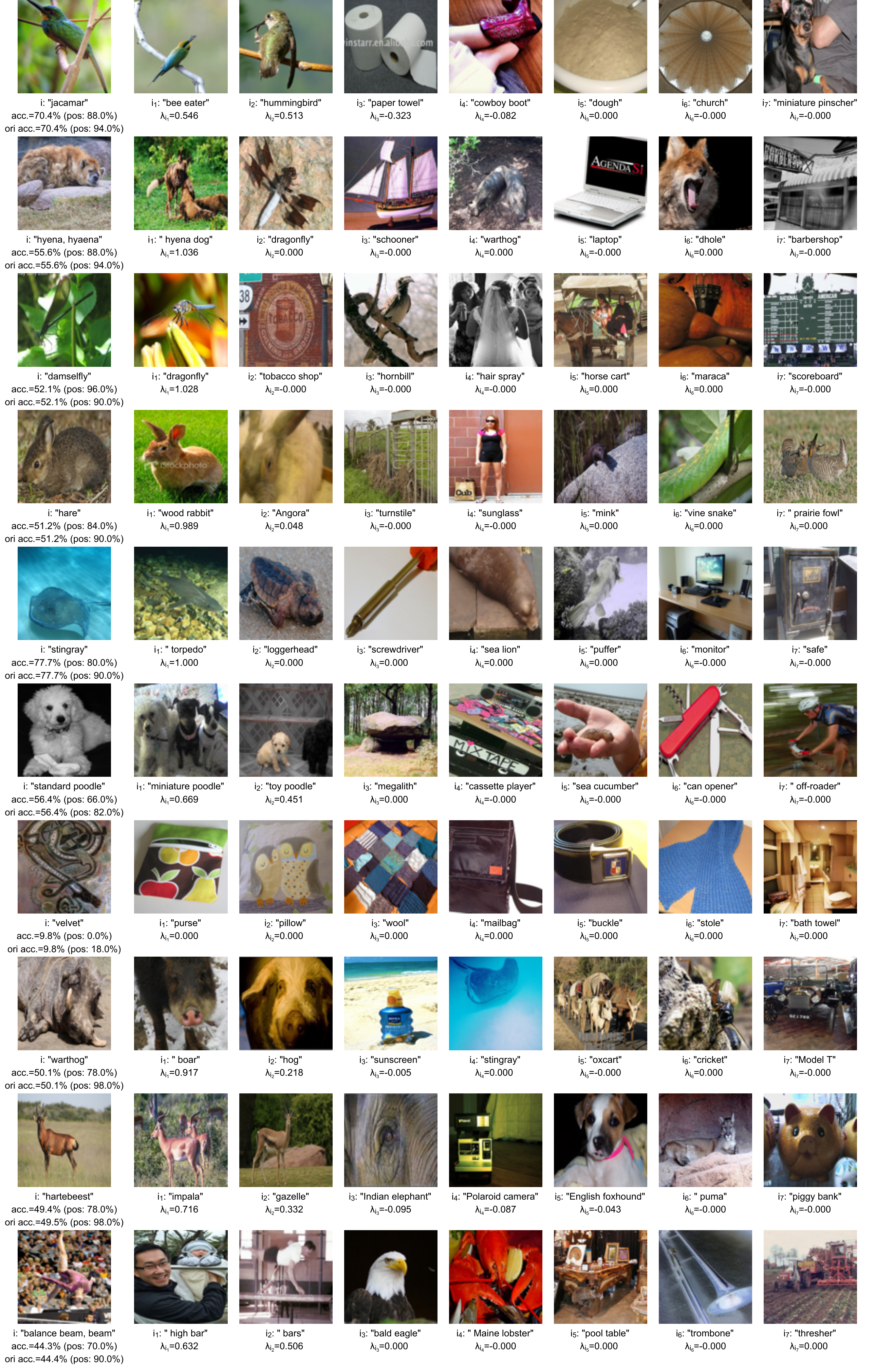}
\caption{Results from ResMLP-S24, where `acc.' and `ori acc.' denote the classification accuracies on the ImageNet validation set, while `pos: xx\%' is the accuracy on positive samples only.}\label{fig:supp_deficit_resmlp}
\end{figure}

\begin{figure}[t]
\centering
\includegraphics[width=0.98\linewidth]{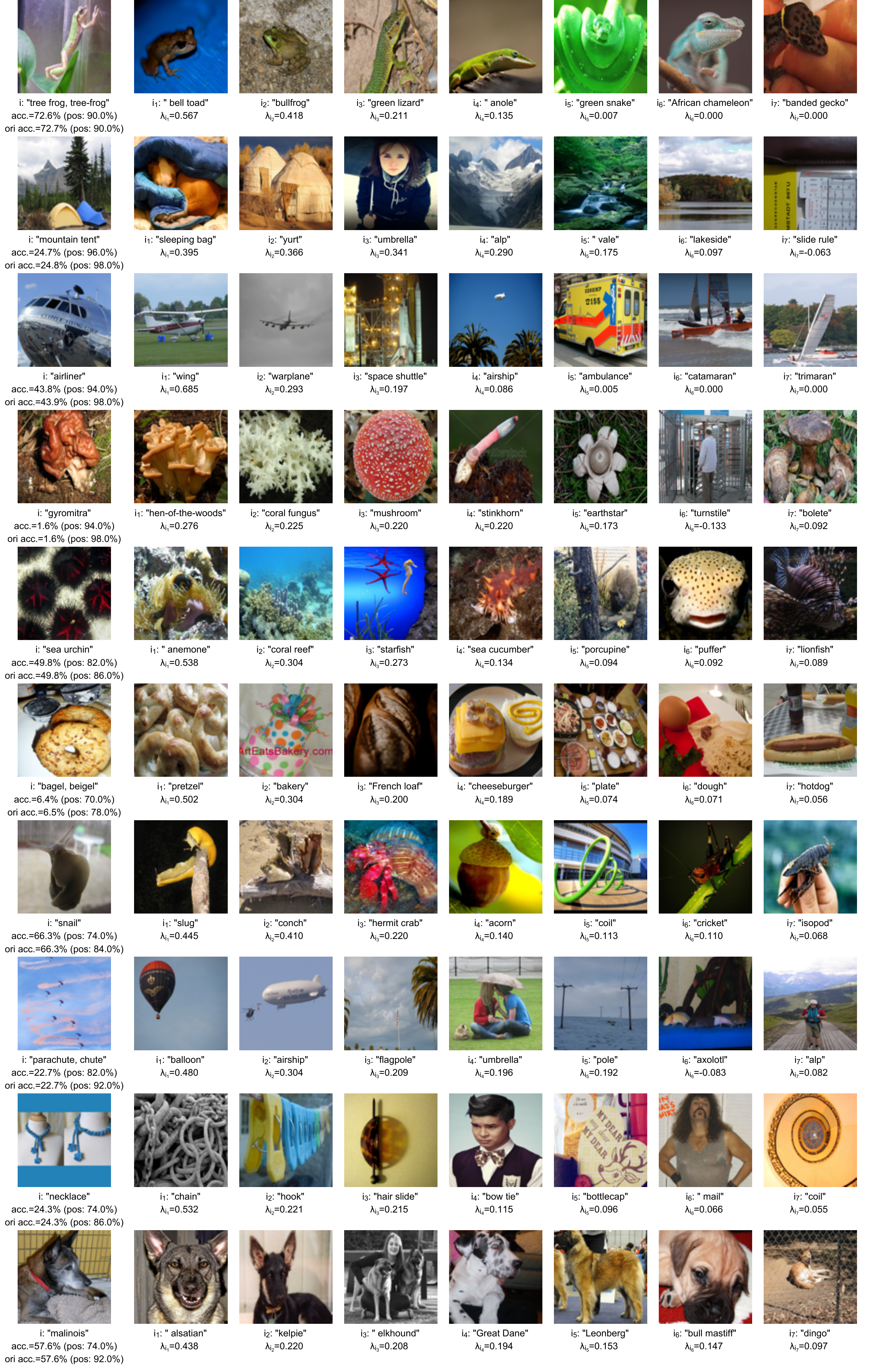}
\caption{Results from ViT-T, where `acc.' and `ori acc.' denote the classification accuracies on the ImageNet validation set, while `pos: xx\%' is the accuracy on positive samples only.}\label{fig:supp_deficit_vit}
\end{figure}

\begin{figure}[t]
\centering
\includegraphics[width=0.98\linewidth]{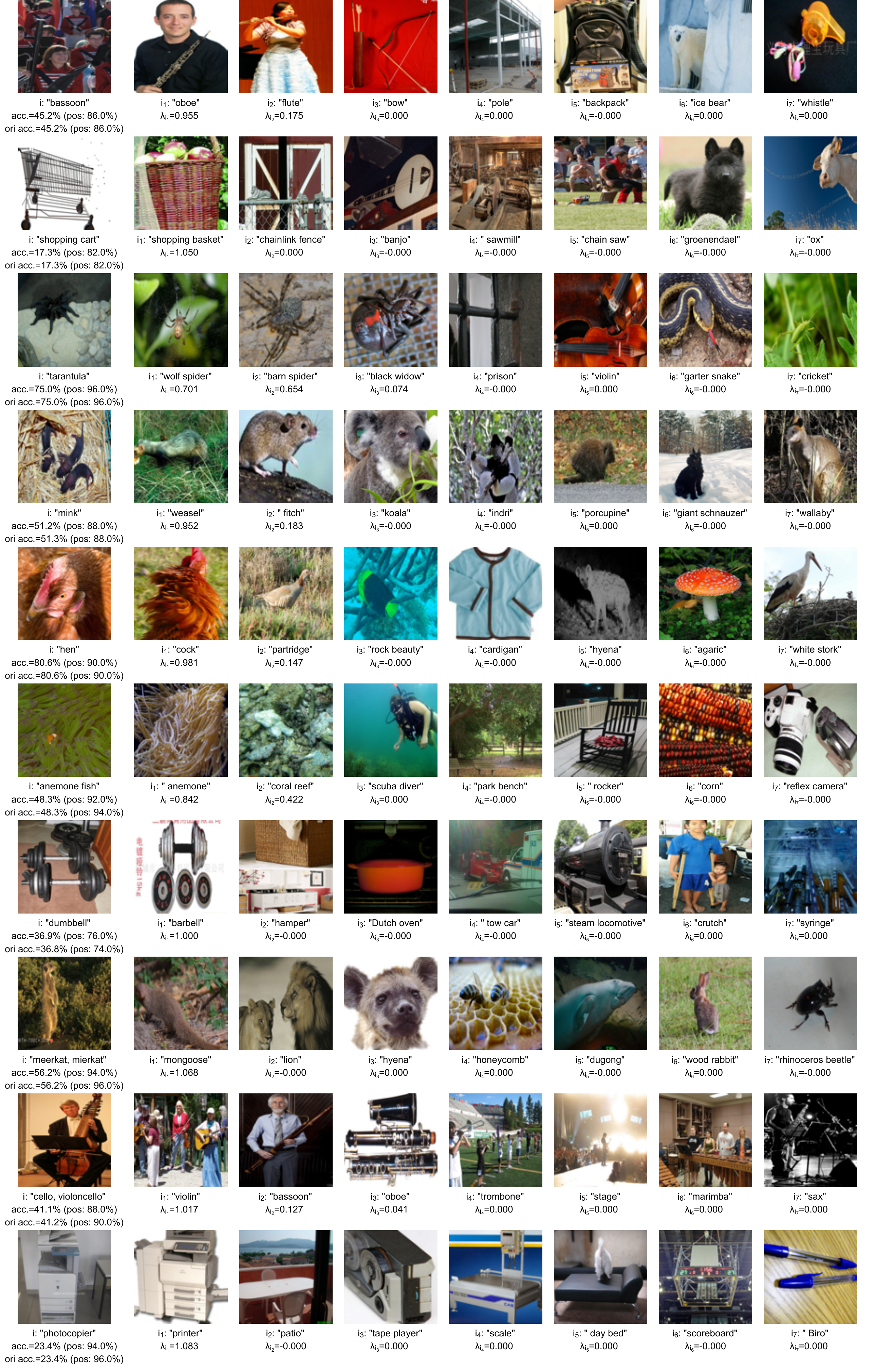}
\caption{Results from Swin-T, where `acc.' and `ori acc.' denote the classification accuracies on the ImageNet validation set, while `pos: xx\%' is the accuracy on positive samples only.}\label{fig:supp_deficit_swin}
\end{figure}

We provide more examples of independence deficit, which shows that classification confidence of some categories can be lineally decided by a few other categories with fixed coefficients. The results obtained by ResNet-18, ResNet-50, GluMixer-24, ResMLP-S24, ViT-T, and Swin-T are reported in \cref{fig:supp_deficit_res18} - \cref{fig:supp_deficit_swin}, respectively. All the results are obtained by solving the Lasso problem in \cref{sec:discussion}. In each figure, we report the classification accuracy for category `$i$': the accuracy by calculating logits with \cref{eq:hamster} is reported as `acc.'; and the original model accuracy is reported as `ori. acc.'. Both the metrics are measured in the whole ImageNet validation set. We further report the classification accuracy on positive samples only for both metrics as `pos' following `acc.' and `ori. acc.' correspondingly. The results show a universal independence deficit phenomenon for broad categories in all those deep networks.




\clearpage
\begin{algorithm}[t]
\caption{Pseudocode of Partial Rank of the Jacobian.}
\label{alg:PartialRank}
\algcomment{\fontsize{7.2pt}{0em}\selectfont
\texttt{matmul}: matrix multiplication;
\texttt{jacobian}: calculate the jacobian matrix;
\texttt{matrix\_rank}: calculate the numerical rank of matrix.
}
\definecolor{codeblue}{rgb}{0.25,0.5,0.5}
\lstset{
  backgroundcolor=\color{white},
  basicstyle=\fontsize{7.2pt}{7.2pt}\ttfamily\selectfont,
  columns=fullflexible,
  breaklines=true,
  captionpos=b,
  commentstyle=\fontsize{7.2pt}{7.2pt}\color{codeblue},
  keywordstyle=\fontsize{7.2pt}{7.2pt},
}
\begin{lstlisting}[language=python]
# image: input images
# model: network
# row_idx, col_idx, patch_size: select a patch of the image to calculate Jacobian matrix

from functools import partial
import torch.nn.functional as functional

def Jacobian_rank(image, model):
    # select a patch of the image to calculate Jacobian matrix
    assert image.size(2) == image.size(3)
    image_size = image.size(2)
    image = image[:, :, row_idx:row_idx + patch_size, col_idx:col_idx + patch_size]
    zero_pad = partial(functional.pad, pad=[(image_size - patch_size) // 2 for _ in range(4)], value=0.)

    # calculate the jacobian matrix
    jacobian_matrix = jacobian(partial(net.forward, preprocess=zero_pad), image)

    # adopt trick to predict the singular values
    jacob = jacob.view(-1, image.size(1) * patch_size * patch_size)
    jacob = matmul(jacob.T, jacob)

    # calculate the partial rank of Jacobian matrix
    return matrix_rank(jacob, symmetric= True)

\end{lstlisting}
\end{algorithm}

\begin{algorithm}[t]
\caption{Pseudocode of Perturbed PCA Dimension of Feature Spaces.}
\label{alg:NumericalRank}
\algcomment{
\fontsize{7.2pt}{0em}\selectfont
\texttt{matmul}: matrix multiplication;
\texttt{randn\_like}: sample a random tensor from a Gaussian distribution;
\texttt{matrix\_rank}: calculate the numerical rank of matrix.
}
\definecolor{codeblue}{rgb}{0.25,0.5,0.5}
\lstset{
  backgroundcolor=\color{white},
  basicstyle=\fontsize{7.2pt}{7.2pt}\ttfamily\selectfont,
  columns=fullflexible,
  breaklines=true,
  captionpos=b,
  commentstyle=\fontsize{7.2pt}{7.2pt}\color{codeblue},
  keywordstyle=\fontsize{7.2pt}{7.2pt},
}
\begin{lstlisting}[language=python]
# image: input images
# model: network
# mag_perturb: magnitude of perturbations
# n_perturb: number of perturbations

def Perturbed_Dimension(image, model, mag_perturb=1e-3, n_perturb=5000):
    # extract features with random perturbations
    features = []
    for _ in range(n_perturb):
        # sample random perturbation from Gaussian distribution
        perturb = randn_like(image) * mag_perturb
        # extract feature
        feature = model(image + perturb)
        features.append(feature)
    features = concatenation(features, dim=0)

    # calculate the covariance matrix
    x = input- mean(input, dim=0)
    x = x.view(x.size(0), -1)
    cov_matrix = matmul(x.T, x) # covariance matrix

    # calculate the perturbed PCA dimensions
    return matrix_rank(cov_matrix, symmetric= True)

\end{lstlisting}
\end{algorithm}

\begin{algorithm}[t]
\caption{Pseudocode of the Classification Dimension of the Final Feature Manifold.}
\label{alg:IntrinsicDimension}
\algcomment{\fontsize{7.2pt}{0em}\selectfont
\texttt{matmul}: matrix multiplication;
\texttt{evd}: eigen value decomposition;
\texttt{calc\_acc}: calculating classification accuracy.
}
\definecolor{codeblue}{rgb}{0.25,0.5,0.5}
\lstset{
  backgroundcolor=\color{white},
  basicstyle=\fontsize{7.2pt}{7.2pt}\ttfamily\selectfont,
  columns=fullflexible,
  breaklines=true,
  captionpos=b,
  commentstyle=\fontsize{7.2pt}{7.2pt}\color{codeblue},
  keywordstyle=\fontsize{7.2pt}{7.2pt},
}
\begin{lstlisting}[language=python]
# image: input images
# model: network
# target: ground-truth labels
# acc_ratio: threshold for measuring intrinsic dimensions of final features

def PCA(X, n_components):
    n = X.shape[0]
    X_mean = mean(X, dim=0, keepdim=True)
    X = X - X_mean
    covariance_matrix = 1 / n * matmul(X.T, X)
    eigenvalues, eigenvectors = evd(covariance_matrix, eigenvectors=True)
    eigenvalues = norm(eigenvalues, dim=1)  # modulus of complex numbers
    idx = argsort(-eigenvalues)
    eigenvectors = eigenvectors[:, idx]
    eigenvectors = eigenvectors[:, :n_components]
    return eigenvectors

def Feature_projection(X, V):
    X_proj = zeros_like(X)
    for component_idx in range(V.size(1)):
        eig_vec = V[:, component_idx].unsqueeze(-1)
        eig_vec_norm = eig_vec / norm(eig_vec, p=2, keepdim=True)
        w_proj = matmul(X, eig_vec_norm)
        X_proj_i = w_proj * eig_vec_norm.T
        X_proj += X_proj_i
    return X_proj

def Intrinsic_dimension(image, model, target, acc_ratio=0.95):
    # pre-extract features and calculate original classification accuracy
    feats = model(image) # [n_samples * n_channels]
    acc_ori = calc_acc(feats, target)

    for n_component in range(1, feats.size(1)):
        # compute the eigenvalues and eigenvectors of a real square matrix
        components = PCA(feats, n_component) # [n_channels * n_component]

        # reconstruct features with principal components
        feats_rec = Feature_projection(feats, components)

        # calculate classification accuracy
        acc = calc_acc(feats_rec, target)

        # return classification dimension
        if acc >= acc_ratio * acc_ori:
            return n_component

\end{lstlisting}
\end{algorithm}

\begin{algorithm}[t]
\caption{Pseudocode of Independence Deficit.}
\label{alg:Hamster}
\algcomment{
\fontsize{7.2pt}{0em}\selectfont \texttt{matmul}: matrix multiplication;
\fontsize{7.2pt}{0em}\selectfont \texttt{mse}: mean squared error;
\fontsize{7.2pt}{0em}\selectfont \texttt{l1\_norm}: sum of the magnitudes of the vectors in a space.
}

\definecolor{codeblue}{rgb}{0.25,0.5,0.5}
\lstset{
  backgroundcolor=\color{white},
  basicstyle=\fontsize{7.2pt}{7.2pt}\ttfamily\selectfont,
  columns=fullflexible,
  breaklines=true,
  captionpos=b,
  commentstyle=\fontsize{7.2pt}{7.2pt}\color{codeblue},
  keywordstyle=\fontsize{7.2pt}{7.2pt},
}
\begin{lstlisting}[language=python]
# image: input images
# model: network
# target2index: dictionary mapping from category index to sample indices
# lr: learning rate for Lasso optimization
# n_iteration: number of iterations for Lasso optimization
# w_reg: weight of the L1 regularization term

def Feature_split(feats, class_i, target2index):
    sample_indices = target2index[class_i]
    start_idx, end_idx = sample_indices[0], sample_indices[-1]
    feats_i = feats[start_idx:end_idx+1, :]
    feats_i_n = concatenation((feats_i[:, :class_i], feats_i[:, class_i+1:]), dim=1)
    feats_i_p = feats_i[:, class_i:class_i+1]
    return feats_i_n, feats_i_p

def Independence_deficit(image, model, target2index, lr=1e-5, n_iteration=5000, w_reg=20.0):
    # pre-extract logits
    logits = model(image) # [n_samples * n_classes]

    # Lasso optimization
    for class_i in range(logits.size(1)):
        # split features by category index
        feats_n, feats_p = Feature_split(logits, class_i, target2index)

        # initialize the linear coefficients of category i
        param = Parameter(zeros(feats_n.size(1), 1))

        # start training
        for _ in range(n_iteration):
            loss = mse(matmul(feat_n, param), feat_p) + w_reg * l1_norm(param)
            loss.backward()
            param -= lr * param.grad

        # save trained coefficients
        save(param)

\end{lstlisting}
\end{algorithm}




\clearpage

\end{document}